\documentclass[lettersize, journal]{IEEEtran}
\IEEEoverridecommandlockouts
\usepackage{cite}
\usepackage{amsmath,amssymb,amsfonts}
\usepackage{algorithm}
\usepackage{algpseudocode}
\usepackage{graphicx}
\usepackage{textcomp}
\usepackage{xcolor}
\usepackage{comment}
\def\BibTeX{{\rm B\kern-.05em{\sc i\kern-.025em b}\kern-.08em
    T\kern-.1667em\lower.7ex\hbox{E}\kern-.125emX}}

\usepackage{bm}
\usepackage[caption=false]{subfig}
\newtheorem{theorem}{Theorem}

\usepackage{amsmath}

\DeclareMathOperator*{\argmin}{arg\,min}

\usepackage{tabularx}

\begin{document}



\title{Age-Based Scheduling for Mobile Edge Computing: A Deep Reinforcement Learning Approach}

\author{
  Xingqiu He,~\IEEEmembership{Member, IEEE},
  Chaoqun You,~\IEEEmembership{Member,~IEEE},
  Tony Q. S. Quek,~\IEEEmembership{Fellow,~IEEE}
  \thanks{X. He and C. You are with the Intelligent Networking and Computing Research Center and School of Computer Science, Fudan University, Shanghai, China,
    (emails: hexqiu@gmail.com, chaoqunyou@gmail.com).
  }
  \thanks{T. Q. S. Quek is with the Singapore University of Technology and Design, Singapore 487372, 
  and also with the Yonsei Frontier Lab, Yonsei University, South Korea (e-mail: tonyquek@sutd.edu.sg).}
  \thanks{This paper is supported in part by the National Research Foundation, Singapore and Infocomm Media Development Authority 
  under its Future Communications Research \& Development Programme.}
  \thanks{Corresponding authors: Chaoqun You and Tony Q. S. Quek.}
}



\maketitle

\begin{abstract}
    With the rapid development of Mobile Edge Computing (MEC), various real-time applications have been deployed to benefit people's daily lives. 
    The performance of these applications relies heavily on the freshness of collected environmental information, which can be quantified by its Age of Information (AoI). 
    In the traditional definition of AoI, it is assumed that the status information can be actively sampled and directly used. 
    However, for many MEC-enabled applications, the desired status information is updated in an event-driven manner and necessitates data processing. 
    To better serve these applications, we propose a new definition of AoI and, based on the redefined AoI, we formulate an online AoI minimization problem for MEC systems. 
    Notably, the problem can be interpreted as a Markov Decision Process (MDP), thus enabling its solution through Reinforcement Learning (RL) algorithms. 
    Nevertheless, the traditional RL algorithms are designed for MDPs with completely unknown system dynamics and hence usually suffer long convergence times. 
    To accelerate the learning process, we introduce Post-Decision States (PDSs) to exploit the partial knowledge of the system's dynamics. 
    We also combine PDSs with deep RL to further improve the algorithm's applicability, scalability, and robustness.
    Numerical results demonstrate that our algorithm outperforms the benchmarks under various scenarios.
\end{abstract}


\section{Introduction} \label{section:introduction}
In recent years, the development of Internet of Things (IoT) technology and the rapid proliferation of wireless devices (WDs)
have given rise to many real-time applications, such as anomaly detection in sensor networks \cite{erhan2021smart} and 
intelligent surveillance \cite{ko2018deep}.
In these applications, the collected raw data (e.g. pictures and videos) usually require further processing before the embedded information is revealed.
For example, videos captured by surveillance cameras need to be analyzed by deep learning algorithms
to determine if an intrusion has occurred.
However, constrained by physical size and manufacturing costs,
the ubiquitously deployed WDs are usually unable to provide satisfying computational capability.
To address this issue, Mobile Edge Computing (MEC) \cite{mao2017survey, mach2017mobile} has been proposed as a new computing paradigm
that enables WDs to offload workload to nearby edge servers through wireless channels.
Compared to traditional cloud computing,
MEC significantly reduces the computation latency and thus substantially improves the Quality of Experience (QoE).

Since the emergence of MEC, extensive works\footnote{Please see \cite{mao2017survey, mach2017mobile} and the references therein.} 
have been conducted to study the computation offloading and task scheduling algorithms
that aim to minimize the latency of tasks or maximize the utilization of system resources.
However, as pointed out in \cite{kaul2012real}, optimizing these performance metrics does not necessarily mean
optimizing the timeliness of information even in the simplest queueing systems.
Therefore, for real-time applications that require the freshest environmental information,
the conventional latency-oriented or utilization-oriented algorithms may result in suboptimal performance.

To solve this problem, the Age of Information (AoI) has recently been proposed as a new metric to measure the 
freshness of information \cite{kaul2011minimizing, kaul2012real, kosta2017age}.
The authors assume that the status information at the sources can be actively sampled
and directly used without further processing \cite{yates2015lazy}.
Hence, the AoI is simply defined as the time elapsed since the generation of the last received status update.
However, in many MEC-enabled applications, the status information is updated in an event-driven manner
(i.e. a new update is sent only if the status changes)
and data processing is required to extract the desired information.
For example, in anomaly detection,
the data collected by sensors are transmitted to the edge server for analysis only when the monitoring indices exceed normal ranges.
To better serve these applications, we revise the classical AoI concept
to accommodate the event-driven sampling policy and the additional processing time.
Specifically, in this paper, the AoI is defined as the time elapsed since the generation of the earliest unprocessed data packet.

Based on the new definition of AoI, we formulate the AoI minimization problem for MEC-enabled applications under the constraints of limited spectrum bandwidth and energy consumption. 
Compared to the latency of tasks, AoI exhibits more complex patterns and is more challenging to optimize \cite{kaul2012real}. 
Moreover, in our problem, the acquisition of status information necessitates data processing.
Therefore, minimizing the redefined AoI requires a joint optimization of the data transmission and data processing processes, 
which further increases the complexity of problem-solving.
To address these issues, we transform the formulated optimization problem into an equivalent Constrained Markov Decision Process (CMDP)
and ameliorate the conventional Reinforcement Learning (RL) algorithms to efficiently learn the optimal control policy.
Our main contributions are summarized as follows.
\begin{enumerate}
    \item We extend the conventional AoI definition and formulate the AoI minimization problem 
        for MEC-enabled applications in an online setting.
        In order to solve the formulated problem, we construct an equivalent CMDP problem and relax its time-average energy consumption constraint via the Lagrangian method.
        We show that the optimal control policy can be obtained by applying RL algorithms (e.g. Q-learning) to the relaxed problem.
    \item The conventional RL algorithms are designed for MDPs with completely unknown system dynamics and hence usually suffer a long convergence time.
        However, in our problem, we have partial knowledge about the system dynamics.
        To improve the learning efficiency, we introduce Post-Decision States (PDSs) to split the known and unknown system dynamics
        so that we only need to learn the unknown part.
        As a result, the algorithm's convergence speed is substantially improved.
    \item To further enhance the algorithm's applicability, scalability, and robustness, we combine PDSs with 
      Deep Deterministic Policy Gradient (DDPG), a classical Deep Reinforcement Learning (DRL) algorithm,
      to obtain a deep PDS learning algorithm.
      We also present several techniques to improve the learning efficiency and stability of the proposed algorithm, including
      redesign of cost functions and normalization of decision variables.
    \item We conduct extensive simulations to validate the performance of our algorithm.
        Numerical results demonstrate that our algorithm is highly efficient and outperforms the benchmarks under various scenarios.
        Our source code is available on Github\footnote{https://github.com/XingqiuHe/DPDS}, allowing researchers to reproduce our experiments and build upon our work.
\end{enumerate}

The rest of the paper is organized as follows.
In Section \ref{section:related_work}, we review related works.
In Section \ref{section:system_model}, we describe the system model and formulate the AoI minimization problem.
The CMDP problem and PDS method are presented in Section \ref{section:algorithm_design}.
After that, we propose the DDPG-based deep PDS learning algorithm in Section \ref{section:dpds}.
In Section \ref{section:simulation}, numerical results are demonstrated to validate the performance of our algorithm.
Section \ref{section:conclusion} concludes the paper and discusses open problems for future work.

\section{Related Work} \label{section:related_work}
Since the first introduction of AoI in \cite{kaul2011minimizing}, many researchers have analyzed the AoI in various systems based on the methodology from queueing theory.
The authors in \cite{kaul2012real} derived analytical expressions for AoI in $M/M/1$, $M/D/1$, and $D/M/1$ systems.
The results were later extended to other queueing models, such as $G/G/1$ \cite{soysal2019age}, $M/G/1$ \cite{sac2018age}, and $D/G/1$ \cite{champati2018statistical}.
Compared to the First-Come-First-Served (FCFS) principle, the authors in \cite{kaul2012status} showed that AoI is improved in Last-Come-First-Served (LCFS) queueing systems.
A variant concept of AoI, named the peak AoI (PAoI), was introduced in \cite{costa2014age},
which provides information about the maximum AoI prior to the reception of each update.
The exact distribution of PAoI in a $PH/PH/1/1$ queue is analyzed in \cite{akar2020finding}
and the authors in \cite{xu2020peak} computed the expected PAoI for a priority queueing system under various scenarios.
In \cite{costa2016age}, the effects of packet management on both AoI and PAoI are analyzed under different pre-emption policies.
Apart from the point-to-point scenario, the AoI in queueing systems with multiple sources or multiple servers is also investigated in \cite{yates2018age, najm2018status, kadota2016minimizing, sun2018age}.
In particular, the authors in \cite{yates2018age} derived a general result for the AoI in multi-source systems under different service principles.
Najm and Telatar \cite{najm2018status} studied the AoI and PAoI in a $M/G/1/1$ model
where the preemptive queue is shared by multiple streams of updates.
They found that one can prioritize a stream from an age point of view by simply increasing its generation rate.
The authors in \cite{kadota2016minimizing} studied the scheduling policy that minimizes AoI in a wireless network with multiple destinations and unreliable channels.
Their model was extended to a multi-source case by Sun \textit{et al.} \cite{sun2018age}
and two (near) optimal scheduling policies are proposed for
any time-dependent, symmetric, and non-decreasing penalty function of the ages.

The aforementioned studies typically assume that update packets sent by information sources can be directly utilized.
However, this assumption may not hold in various MEC applications where data processing is required to extract desirable information from raw data.
To fill the gap, Song \textit{et al.} \cite{song2019age} proposed a new metric called Age of Task (AoT), which incorporates the computation time in the conventional AoI concept.
By jointly considering the task scheduling and computation offloading,
the authors proposed a lightweight task offline algorithm to optimize the AoT under the energy consumption constraint.
A similar metric, named Age of Processing (AoP), is also presented in \cite{li2021age} where the authors utilize the Lagrangian transformation framework to solve the formulated CMDP problem.
The concept of AoP is also employed in \cite{ndikumana2022age} and \cite{ndikumana2023age} to facilitate the selection of multi-radio access technologies
and allocation of communication and computational resources.
The authors in \cite{zou2021optimizing} and \cite{xu2019optimizing} investigated the AoI in MEC systems where the collected data can be (partially) pre-processed at the sensors.
Considering task-specific timeliness requirements of different applications, a low-complexity scheduling policy based on Lyapunov optimization 
is presented in \cite{sun2023optimizing}.
The authors also analyzed the performance gap between the proposed policy and a theoretical lower bound of the time-average AoI penalty.

Recently, there has been extensive interest in applying DRL methods for resource management in communication systems 
\cite{naderializadeh2021resource, xiong2020resource, ju2023joint, zhao2019deep, ye2019deep}.
Among these, the works most closely related to ours are \cite{peng2023aoi, jiang2023age, chen2023joint, xie2023minimizing},
as they also focus on minimizing AoI in MEC networks.
Specifically, in \cite{peng2023aoi}, the authors considered task dependencies and proposed an offloading algorithm based on an improved dueling double deep Q-network.
Jiang \textit{et al.} \cite{jiang2023age} established a joint optimization model for computation offloading and transmission scheduling.
To address the strong coupling between these processes, the original problem is decomposed into two stages, with a DRL algorithm proposed for each stage.
Chen \textit{et al.} \cite{chen2023joint} investigated the joint optimization of sensing and computation for MEC-enabled IoT networks with limited computational resources.
In \cite{xie2023minimizing}, a new metric, Age of Usage Information (AoUI), is introduced to jointly capture the 
freshness and usability of correlated data in IoT networks.
After that, the authors employed double deep Q-network (DDQN) to construct the optimal data scheduling algorithm that minimizes AoUI.
It's worth noting that the DRL algorithms used in these works are originally designed for MDPs with completely unknown system dynamics. 
However, in many communication systems, part of the system dynamics is known in advance, 
hence directly applying these DRL algorithms would result in suboptimal results.
To address this issue, this paper enhances existing DRL algorithms significantly by introducing PDSs to leverage known information about system dynamics. 
Experimental results demonstrate that our approach substantially outperforms existing DRL algorithms.

\section{System Model and Problem Formulation} \label{section:system_model}
In this section, we first describe the considered MEC system and the computation offloading model.
Then we introduce the concept of AoI, which measures the freshness of information contained in computation-intensive tasks.
At last, we formulate an online optimization problem that aims to minimize the time-average AoI under
limited spectrum bandwidth and energy consumption.
The major notations in this paper are summarized in Table \ref{tab:notation}.

\begin{table}[!t]
\renewcommand{\arraystretch}{1.2}
\caption{Major Notations}
\label{tab:notation}
\centering
\begin{tabularx}{0.99\linewidth}{l l}
\hline
\textbf{Notation} & \textbf{Description}\\
\hline
$N$ & Number of WDs. \\
$T$ & Number of time slots. \\
$\Delta t$ & Duration of each time slot. \\
$\mathcal{J}_i$ & Index set of tasks generated by WD $i$. \\
$\tau_{ij}$ & The $j$-th task generated by WD $i$. \\
$t_{ij}, d_{ij}$ & Generation time and data size of task $\tau_{ij}$. \\
$\kappa$ & Number of CPU cycles required to process one bit of data. \\
$\gamma$ & Energy efficiency factor related to the chip architecture. \\
$f_i(t)$ & CPU frequency of WD $i$ at slot $t$. \\
$P_i(t)$ & Transmission power of WD $i$ at slot $t$. \\
$h_i(t)$ & Channel gain of WD $i$ at slot $t$. \\
$W_i(t)$ & Spectrum bandwidth allocated to WD $i$ at slot $t$. \\
$r_i(t)$ & Transmission rate of WD $i$ at slot $t$. \\
$d^l_i(t)$ & Amount of locally processed data of WD $i$ at slot $t$. \\
$d^o_i(t)$ & Amount of data transmitted from WD $i$ to the BS at slot $t$. \\
$d^r_i(t)$ & Amount of remaining data in the HOL task of WD $i$. \\
$E^l_i(t)$ & Energy consumption due to local computing. \\
$E^o_i(t)$ & Energy consumption due to computation offloading. \\
$a_i(t)$ & AoI of WD $i$ at slot $t$. \\
$b_i(t)$ & Whether the HOL task of WD $i$ is completed at slot $t$. \\
$g_i(t)$ & Whether a new task is generated by WD $i$ at slot $t$. \\
$q_i(t)$ & Number of queueing tasks in WD $i$ at slot $t$. \\
\hline
\end{tabularx}
\end{table}

\subsection{Network Model}
\begin{figure}[!t]
    \centering
    \includegraphics[width=0.35\textwidth]{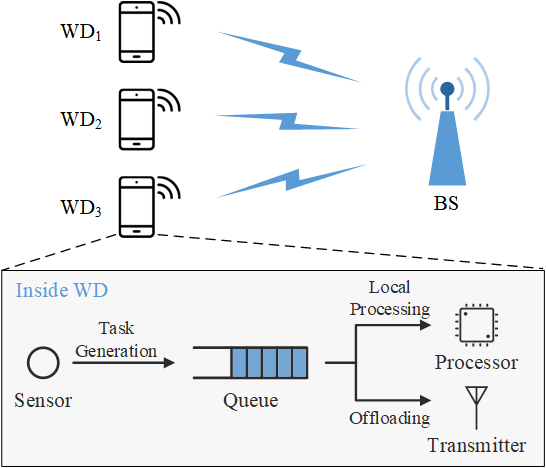}
    \caption{A simple example of the considered system model.}
    \label{fig:system}
\end{figure}
As shown in Fig. \ref{fig:system}, we consider a MEC system consisting of one base station (BS) and $N$ WDs.
We assume the BS is endowed with computational resources and serves as an edge server.
The WDs monitor the surrounding environment and 
generate a computation task containing related raw data when the environment changes.
The environmental information can only be obtained after data processing.
In this paper, we focus on applications that require sequential data processing,
which means the processing of one task relies on the computation results of previous tasks.
One illustrative example is time-series anomaly detection, where the identification of abnormal patterns relies on historical environmental information.
Consequently, the tasks on each WD must be processed in a First-Come-First-Serve (FCFS) manner.
Nevertheless, our approach used to design scheduling algorithms is not bound by the FCFS property and can be easily extended to other applications
where the processing of tasks is mutually independent.

Consider a sufficiently long time interval that is 
divided into $T$ time slots of equal length $\Delta t$. 
Let $\mathcal{N} = \{ 1,2,\dots,N \}$ be the set of WDs and $\mathcal{J}_i = \{ 1,2,\dots,J_i \}$ be the index set of tasks generated by WD $i\in\mathcal{N}$.
For any $j\in\mathcal{J}_i$, the task $\tau_{ij}$ is characterized by a two tuple $\tau_{ij} = (t_{ij}, d_{ij})$,
where $t_{ij}$ is the generation time of $\tau_{ij}$ and $d_{ij}$ is the task's data size (in bits). 
As in \cite{song2019age}, we assume each task can be further partitioned into smaller subtasks,
and each subtask can be either processed by the local processor or offloaded to the BS via wireless transmission.
This assumption is known as the ``partial offloading'' and is widely adopted in works studying the computation offloading in MEC systems \cite{wang2017joint, kuang2019partial, wang2016mobile}.
No matter where the task is processed, the computation result will be synchronized between WDs and the BS so that the next task can be handled in both places.
Since the data size of the computation results is usually very small, the synchronization time and corresponding energy consumption are negligible.
Next, we will present the computation model of local processing and computation offloading.

\subsubsection{Local processing}
Due to the limited battery capacity of WDs, each WD should allocate its energy reasonably for local processing
and data offloading.
To control the energy consumption of local processing,
we use Dynamic Voltage and Frequency Scaling (DVFS) \cite{rabaey2003digital}
to dynamically adjust the working frequencies of WDs' processors.
Let $f_i(t)$ be the CPU frequency of WD $i$ at slot $t$, $\kappa$ is the number of CPU cycles required to process one bit of data,
then the amount of locally-processed data at slot $t$ is
\begin{equation}
    d^l_i(t) = \frac{f_i(t) \Delta t}{\kappa}.
    \label{eq:d^l_i}
\end{equation}
According to the results in \cite{chandrakasan1992low}, the energy consumption per CPU cycle is proportional to the square of the frequency,
which can be expressed as $\gamma f^2_i(t)$, where $\gamma$ is the energy-efficiency factor related to the chip architecture.
Combining with \eqref{eq:d^l_i}, the energy consumption due to local computation is given by
\begin{align}
    E^l_i(t) &= \gamma f^2_i(t) \times f_i(t) \Delta t = \frac{\gamma \kappa^3}{\Delta t^2} \left( d^l_i(t) \right)^3.
    \label{eq:E^l_i}
\end{align}

\subsubsection{Computation offloading}
In this paper, we assume the Orthogonal Frequency-Division Multiple Access (OFDMA) technology is adopted to avoid mutual interference among WDs.
According to the Shannon's formula, the transmission rate of WD $i$ can be formulated as
\begin{equation}
    r_i(t) = W_i(t) \log_2 \left( 1 + \frac{P_i(t) h_i(t)}{\sigma^2} \right),
    \label{eq:r_i}
\end{equation}
where $W_i(t)$ is the spectrum bandwidth allocated to WD $i$ at slot $t$,
$P_i(t)$ is the transmission power of WD $i$,
$h_i(t)$ is the channel gain between WD $i$ and the BS,
and $\sigma^2$ denotes the noise power at the BS.
All channels follow quasi-static flat-fading \cite{mao2019energy, wu2019online, wang2020optimal}, i.e. the channel gain remains constant during each time slot, 
but may vary at the boundaries of slots.
Hence, the amount of data offloaded from WD $i$ to the BS at slot $t$ is
\begin{equation}
    d^o_i(t) = r_i(t) \Delta t
    \label{eq:d^o_i}
\end{equation}
and the corresponding energy consumption of WD $i$ is
\begin{equation}
    E^o_i(t) = P_i(t) \Delta t. \label{eq:offloading_energy}
\end{equation}
By substituting \eqref{eq:r_i} and \eqref{eq:d^o_i} into \eqref{eq:offloading_energy} we have
\begin{equation}
    E^o_i(t) = \left( 2^{\frac{d^o_i(t)}{W_i(t)\Delta t}} - 1 \right) \frac{\sigma^2 \Delta t}{h_i(t)}.
    \label{eq:E^o_i}
\end{equation}
We assume the computational capability of the BS is much stronger than that of WDs.
Hence, the computation data offloaded to the BS can be processed within one time slot.

In practice, the duration of each time slot is relatively short so the processing of each task usually spans over multiple slots.
For the convenience of analysis, we require that at most one task is processed on each WD at each time slot.
In other words, if a task is completed in the middle of slot $t$, we will defer the processing of the next task to slot $t+1$.
This means the totally processed data at each slot should not exceed the remaining data in the head-of-line (HOL) task\footnote{
In our problem, HOL task refers to the task that is currently being processed.}.
Let $d^r_i(t)$ be the amount of remaining data in the HOL task of WD $i$ at slot $t$, then $d^l_i(t)$ and $d^o_i(t)$ should satisfy
\begin{equation}
    d^l_i(t) + d^o_i(t) \leq d^r_i(t), \quad \forall i\in\mathcal{N}. \label{cons:remaining_data}
\end{equation}
Notice that constraint \eqref{cons:remaining_data} is purely for the sake of convenience in describing the system model.
We will provide a more detailed explanation in Section \ref{subsection:lagrangian}.
Additionally, to guarantee that the lifetimes of WDs are sufficiently long, their time-average energy consumption should be lower than a threshold $E^{max}_i$.
Let $E_i(t) = E^l_i(t) + E^o_i(t)$ be the energy consumption of WD $i$ at slot $t$, then
this requirement can be expressed by the following constraint
\begin{equation}
    \bar{E}_i \leq E^{max}_i, \quad \forall i\in\mathcal{N}, \label{cons:energy}
\end{equation}
where $\bar{E}_i = \lim_{T\to\infty} 1/T \sum_{t=0}^{T-1} E_i(t)$ is the time-average energy consumption of WD $i$.

\subsection{AoI for MEC-Enabled Applications} \label{subsection:aoi}
As discussed in Section \ref{section:introduction},
the environmental information of many MEC-enabled applications is updated in an event-driven manner
and necessitates data processing, so we need to extend the traditional AoI concept to 
accommodate the event-driven sampling policy and the additional processing time.
However, our algorithm does not rely on the specific AoI expression and thus can be equally applied to other variants of AoI.


Suppose the first task in the system is generated at the initial time slot (i.e. $t=0$).
The instantaneous AoI of WD $i$ at slot $t$, denoted by $a_i(t)$, is defined as the time elapsed since the earliest unprocessed task is generated.
Let $t_{ij}'$ be the finish time of task $\tau_{ij}$,
then the expression of $a_i(t)$ is given by
\begin{equation}
    a_i(t) = t - u_i(t),
\end{equation}
where $u_i(t) = \min \{ t_{ij} : t_{ij} \leq t, t_{ij}' \geq t, \forall j\in\mathcal{J}_i \}$ is the generation time of the HOL task in WD $i$.
If the queue is empty, we set $u_i(t) = t$ and $a_i(t)$ becomes $0$.

\begin{figure}[!t]
    \centering
    \includegraphics[width=0.35\textwidth]{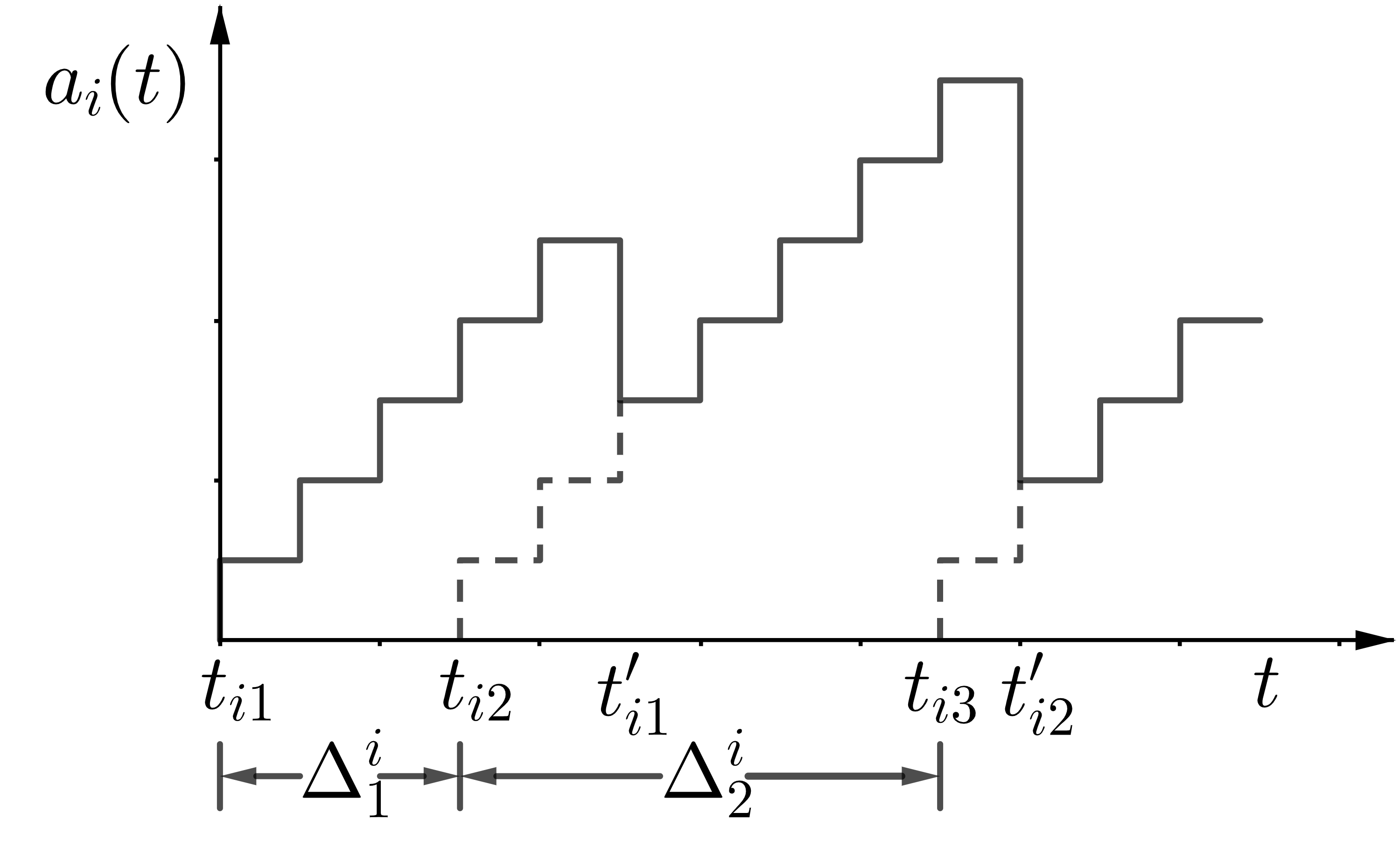}
    \caption{Evolution trace of $a_i(t)$.}
    \label{fig:aoi}
\end{figure}
Fig. \ref{fig:aoi} demonstrates the evolution trace of $a_i(t)$.
If no task of WD $i$ is completed at slot $t-1$, then $a_i(t)$ grows linearly with respect to $t$, so we have $a_i(t) = a_i(t-1) + 1$.
Otherwise, if task $\tau_{ij}$ is completed at slot $t-1$, then $a_i(t)$ will jump downward to $t-t_{i,j+1}$ or $0$, depending on whether $\tau_{i,j+1}$ has been generated.
Let $b_i(t) \in \{ 0,1 \}$ be the binary variable indicating whether the HOL task on WD $i$ is completed at slot $t$,
then the expression of $b_i(t)$ is
\begin{equation}
    b_i(t) = 
    \begin{cases}
        1, &\text{$q_i(t) > 0$ and $d^l_i(t) + d^o_i(t) = d^r_i(t)$} \\
        0, &\text{otherwise}
    \end{cases},
\end{equation}
where $q_i(t)$ denotes the number of queueing tasks on WD $i$ at slot $t$.
The update formula of $q_i(t)$ is 
\begin{equation}
    q_i(t+1) = q_i(t) - b_i(t) + g_i(t),
    \label{eq:update_q}
\end{equation}
where $g_i(t) \in \{ 0,1 \}$ is the binary variable indicating whether a new task is generated by WD $i$ at slot $t$.


\subsection{Problem Formulation}
To improve the freshness of collected status information,
we aim to minimize the long-term AoI of all WDs under limited energy consumption and wireless bandwidth.
With decision variables $f_i(t), P_i(t), W_i(t)$, the AoI minimization problem can be formulated as follows
\begin{align}
  \min_{f_i(t), P_i(t), W_i(t)}\quad & \lim_{T\to\infty} \frac{1}{T} \sum_{t=0}^{T-1} \sum_{i=1}^{N} a_i(t) \label{amp} \\
    s.t.\quad & \eqref{cons:remaining_data}, \eqref{cons:energy} \quad \tag{\ref{amp}{a}} \label{amp:data_energy_cons} \\
    & \sum_{i\in\mathcal{N}} W_i(t) \leq W^{max}, \quad \forall t\in\mathcal{T} \tag{\ref{amp}{b}} \label{amp:bandwidth} \\
    & 0 \leq f_i(t) \leq f^{max}_i, \quad \forall i\in\mathcal{N}, t\in\mathcal{T} \tag{\ref{amp}{c}} \label{amp:variable_f} \\
    & 0 \leq P_i(t) \leq P^{max}_i, \quad \forall i\in\mathcal{N}, t\in\mathcal{T} \tag{\ref{amp}{d}} \label{amp:variable_P} \\
    & 0 \leq W_i(t), \quad \forall i\in\mathcal{N}, t\in\mathcal{T} \tag{\ref{amp}{e}} \label{amp:variable_W}
\end{align}
Constraint \eqref{amp:data_energy_cons} contains the data and energy constraints described above.
Constraint \eqref{amp:bandwidth} ensures the allocated bandwidth does not exceed the BS's capacity $W^{max}$,
where $\mathcal{T} = \{ 0,1,\dots,T-1 \}$ is the index set of all time slots.
The potential values of all decision variables are defined in constraints \eqref{amp:variable_f}, \eqref{amp:variable_P}, and \eqref{amp:variable_W}.
Compared to the latency of tasks, AoI exhibits more complex patterns so problem \eqref{amp} is much more challenging than the
conventional latency-oriented scheduling problems \cite{kaul2012real}.

\section{Online Scheduling Based on Reinforcement Learning} \label{section:algorithm_design}
This section proposes an online scheduling algorithm based on reinforcement learning.
Specifically, we first transform the AoI minimization problem into an equivalent CMDP
and then reformulate it as an unconstrained MDP by introducing Lagrange multipliers associated with 
energy consumption constraints.
The resulting MDP can be solved by classical tabular reinforcement learning algorithms such as Q-learning \cite{watkins1992q}.
However, these algorithms are designed for MDPs with completely unknown system dynamics and hence usually
suffer long convergence times \cite{jaksch2010near, even2004learning}.
To accelerate the learning process,
we leverage the concept of PDSs \cite{salodkar2008line,mastronarde2012joint} to exploit partial knowledge of the system dynamics.
Although our algorithm is much more efficient than the conventional Q-learning,
it still bears several issues that are inherent in tabular reinforcement learning.
To address these issues, we will utilize Deep Neural Networks (DNNs) to further improve our algorithm in the next section.

\subsection{CMDP Formulation} \label{subsection:cmdp_formulation}
We first show that the optimization problem \eqref{amp} is equivalent to a CMDP.
Typically, an MDP is represented by a four tuple $(\mathcal{A}, \mathcal{S}, p, R)$,
where $\mathcal{A}$ and $\mathcal{S}$ are the set of all possible actions and states,
$p$ specifies the state transition probabilities,
and $R$ gives the reward in each step.
In our problem, the reward is defined as the negative of the total AoI in each slot, i.e. $R(t) = - \sum_{i=1}^N a_i(t)$.
The other three components are discussed in the following.

We define the action at slot $t$ as the collection of control decisions, i.e. 
$\alpha(t) = (f_i(t), P_i(t), W_i(t))_{i\in\mathcal{N}} \in \mathcal{A}$.
Then the action space $\mathcal{A}$ is the 
feasible region defined by constraints \eqref{cons:remaining_data} and \eqref{amp:bandwidth}-\eqref{amp:variable_W}. 
We also define the system state as the vector containing the remaining data in HOL task $d^r_i(t)$,
the current AoI $a_i(t)$, the number of queueing tasks $q_i(t)$,
and the channel state $h_i(t)$, i.e. $s(t) = (d^r_i(t), a_i(t), q_i(t), h_i(t))_{i\in\mathcal{N}} \in \mathcal{S}$,
where $\mathcal{S}$ is the set of all possible states.
As suggested in \cite{li2006towards}, solutions of MDPs are usually more efficient if we work in an abstract state space that
only contains important system information.
Therefore, $s(t)$ does not contain other system information such as the time intervals among queueing tasks
because this information is less relevant to the optimization of AoI.

Next, we will discuss the state transition probabilities in our CMDP model.
The values of $d^r_i(t+1)$ and $a_i(t+1)$ depend on the state and action in the previous slot, which can be classified into 
the following cases:
\begin{enumerate}
    \item If $q_i(t) - b_i(t) = 0$, then the task queue is emptied at slot $t$
        and $d^r_i(t+1)$ becomes the data size of the newly generated task
        \begin{align}
            p(&d^r_i(t+1)=d) \notag \\
            &=
            \begin{cases}
                p(g_i(t) = 0), &\text{$d=0$}\\
                p(g_i(t) = 1) \times p(d_{ij} = d), &\text{$d>0$}
            \end{cases}.
            \label{eq:p_d_1}
        \end{align}
        According to the definition of $a_i(t)$, if $g_i(t) = 0$, the task queue is still empty at slot $t+1$
        so we have $a_i(t+1) = 0$.
        Otherwise, if $g_i(t) = 1$, we have $a_i(t+1) = 1$.
        In both cases, $a_i(t+1) = g_i(t)$ so the transition probability is
        \begin{equation}
            p(a_i(t+1) = a) = p(g_i(t) = a), \quad a\in \{0,1\}.
            \label{eq:p_a_1}
        \end{equation}
    \item If $q_i(t) - b_i(t) > 0$ and $b_i(t) = 1$, then $d^r_i(t+1)$ is the data size of the next queueing task
        \begin{equation}
            p(d^r_i(t+1)=d) = p(d_{ij} = d).
            \label{eq:p_d_2}
        \end{equation}
        Since the HOL task is completed, the value of $a_i(t+1)$ jumps downward and we have
        \begin{equation}
            p(a_i(t+1) = a) = p(\Delta_{ij} = a_i(t) + 1 - a),
            \label{eq:p_a_2}
        \end{equation}
        where $\Delta_{ij}$ is the time interval between the arrival time of the HOL task and its successor.
    \item Otherwise, we have $q_i(t) - b_i(t) > 0$ and $b_i(t) = 0$. So the HOL task is not finished yet and thus the values of $d^r_i(t+1)$
        and $a_i(t+1)$ are deterministic
        \begin{align}
            d^r_i(t+1) &= d^r_i(t) - d^l_i(t) - d^o_i(t) \label{eq:p_d_3} \\
            a_i(t+1) &= a_i(t) + 1. \label{eq:p_a_3}
        \end{align}
\end{enumerate}
Similarly, according to the update rule \eqref{eq:update_q}, we can also derive the transition probability of $q_i(t)$ as
\begin{align}
    p(&q_i(t+1) = q) \notag \\
    &=
    \begin{cases}
        p(g_i(t) = 0), &\text{$q = q_i(t) - b_i(t)$} \\
        p(g_i(t) = 1), &\text{$q = q_i(t) - b_i(t) + 1$}.
    \end{cases} 
    \label{eq:p_q}
\end{align}
Hence, for any $s', s\in\mathcal{S}$ and $\alpha\in\mathcal{A}$, the state transition probability is given by
\begin{align*}
    p(s' | s, \alpha) = \prod_{i\in\mathcal{N}} p({d^r_i}', a_i', q_i' | s, \alpha) p(h_i' | h_i),
\end{align*}
where $p({d^r_i}', a_i', q_i' | s, \alpha)$ is defined in \eqref{eq:p_d_1}-\eqref{eq:p_q}.
Notice that we have assumed (i) the update of the channel gain is independent of other components in the system state
and (ii) the state transitions of different WDs are mutually independent.

Let $\pi:\mathcal{S} \to \mathcal{A}$ denote a stationary policy that maps states to actions (i.e. $\alpha(t) = \pi(s(t))$)
and let $\Phi$ be the set of all possible stationary policies.
Starting from the initial state $s$, the time average expectation of total AoI and WD $i$'s energy consumption
under policy $\pi\in\Phi$ are defined as
\begin{equation*}
    \bar{A}^{\pi}(s) = \lim_{T\to\infty} \mathbb{E} \left[ \frac{1}{T} \sum_{t=0}^{T-1} \sum_{i=1}^{N} a_i(s(t), \pi(s(t))) \, | \, s(0) = s \right]
\end{equation*}
and 
\begin{equation*}
    \bar{E}_i^{\pi}(s) = \lim_{T\to\infty} \mathbb{E} \left[ \frac{1}{T} \sum_{t=0}^{T-1} E_i(s(t), \pi(s(t))) \, | \, s(0) = s \right],
\end{equation*}
where we have expressed $a_i(t)$ and $E_i(t)$ as functions of $s(t)$ and $\pi(s(t))$ to indicate their dependence on the system state and action.
As mentioned in \cite{neely2010queue}, the time average expectations are the same as pure time averages (with probability 1) under some mild assumptions.
Therefore, we can use $\bar{A}^{\pi}(s)$ and $\bar{E}_i^{\pi}(s)$ to replace the time average AoI and time average energy consumption defined in \eqref{amp} and \eqref{cons:energy}.
Hence, the AoI minimization problem \eqref{amp} can be transformed into the following equivalent CMDP formulation
\begin{equation}
    \min_{\pi\in\Phi} \bar{A}^{\pi}(s) \quad \mbox{s.t. } \bar{E}_i^{\pi}(s) \leq E^{max}_i, \ \forall i\in\mathcal{N}, s\in\mathcal{S}.
\end{equation}
To solve this optimization problem, we will further reformulate it as an unconstrained MDP via the Lagrangian approach.
The details are presented in the next subsection.

\subsection{The Lagrangian Approach} \label{subsection:lagrangian}
By introducing a Lagrange multiplier $\lambda_i \geq 0$ for the energy consumption constraint of WD $i$,
we can define a Lagrangian cost function as follows
\begin{equation}
    C^{\lambda}(s, \alpha) = \sum_{i\in\mathcal{N}} a_i(s,\alpha) + \sum_{i\in\mathcal{N}} \lambda_i \left( E_i(s,\alpha) - E^{max}_i \right).
    \label{eq:cost_function}
\end{equation}
As stated in the following theorem, with this Lagrangian cost function, we can construct an unconstrained MDP that is equivalent to the CMDP defined above.
\begin{theorem}
    There exist non-negative Lagrangian multipliers $\lambda^* = (\lambda^*_1, \dots, \lambda^*_N) \geq 0$ such that
    the optimal value of the CMDP with initial state $s$ can be computed as
    \begin{equation}
        \bar{A}^{*}(s) = \min_{\pi\in\Phi} V^{\pi, \lambda^*}(s),
    \end{equation}
    where
    \begin{equation*}
        V^{\pi, \lambda}(s) = \lim_{T\to\infty} \mathbb{E} \left[ \frac{1}{T} \sum_{t=0}^{T-1} C^{\lambda}(s(t), \pi(s(t))) \, | \, s(0) = s \right].
    \end{equation*}
    \label{theorem:lagrange}
\end{theorem}
\begin{IEEEproof}
    Please see Chapter 4 and Chapter 12 in \cite{altman1999constrained}.
\end{IEEEproof}

Theorem \ref{theorem:lagrange} 
guarantees that the optimal solutions of the unconstrained MDP with cost function \eqref{eq:cost_function}
are equivalent to that of the CMDP.
Next we introduce several properties and definitions that are useful in solving unconstrained MDPs.
According to the results in \cite{bertsekas2000dynamic}, for any fixed $\lambda$, 
there exists an optimal relative state-value function $V^{*, \lambda}(s)$ and a unique scalar $v^{*,\lambda}$ that satisfy
\begin{equation}
    V^{*, \lambda}(s) + v^{*,\lambda} = \min_{\alpha\in\mathcal{A}} \left\{ C^{\lambda}(s, \alpha) + \sum_{s'\in\mathcal{S}} p(s'|s,\alpha) V^{*, \lambda}(s') \right\}.
\end{equation}
Notice that $V^{*,\lambda}(s)$ is not unique as $V^{*,\lambda}(s)$ plus an arbitrary constant still satisfies the above equality.
Moreover, it can be easily verified that $V^{\pi,\lambda}$ is one of the $V^{*,\lambda}$ when using the optimal policy $\pi^{*, \lambda}$.
Based on $V^{*, \lambda}(s)$, we can also define the optimal relative action-value function $Q^{*, \lambda}(s, \alpha)$ as the one that satisfies
\begin{equation}
    Q^{*, \lambda}(s, \alpha) + v^{*,\lambda} = C^{\lambda}(s, \alpha) + \sum_{s'\in\mathcal{S}} p(s'|s,\alpha) V^{*, \lambda}(s').
\end{equation}
By the definition of $V^{*, \lambda}(s)$, we have 
\begin{equation}
    V^{*, \lambda}(s) = \min_{\alpha\in\mathcal{A}} Q^{*, \lambda}(s, \alpha)
    \label{eq:V_Q}
\end{equation}
and the optimal policy $\pi^{*, \lambda}$ can be computed by
\begin{equation}
    \pi^{*, \lambda}(s) = \argmin_{\alpha\in\mathcal{A}} Q^{*, \lambda}(s, \alpha), \quad \forall s\in\mathcal{S}.
    \label{eq:optimal_policy}
\end{equation}
When the cost function and transition probabilities are known, the optimal policy $\pi^{*, \lambda}$ can be directly computed by using the relative value iteration algorithm \cite{sutton2018reinforcement}.
In our problem, however, the distributions of $g_i(t)$ and $d_{ij}$ are unknown in advance and must be estimated based on experience.
To this end, we adopt a model-free reinforcement learning approach that learns $Q^{*, \lambda}$ and $\pi^{*, \lambda}$ online.

\emph{Remark:} At this stage, we can elaborate on why we need constraint \eqref{cons:remaining_data}.
When describing the state transition in the previous subsection, we have categorized the process into three cases based on 
whether a task was completed in the preceding time slot. 
It's noteworthy that without constraint \eqref{cons:remaining_data}, the system may encounter scenarios where 
(i) multiple tasks are completed in the previous time slot, and (ii) a new task may be partially processed. 
This adds considerable complexity to the state transition description as it introduces numerous new cases where 
the update of $a_i(t)$ and $d^r_i(t)$ exhibits distinct patterns. 
Nevertheless, conducting a comprehensive analysis of the state transition is unnecessary because we cannot directly calculate the transition probabilities due to
unknown distributions of $g_i(t)$ and $d_{ij}$.
Hence, we introduced constraint \eqref{cons:remaining_data} to simplify the analysis of system transition. 
However, we will relax constraint \eqref{cons:remaining_data} in the following algorithm design to achieve more flexible scheduling policies.

\subsection{Inefficiencies of Conventional Q-Learning} \label{subsection:q_learning}
The conventional Q-learning was proposed in \cite{watkins1992q} to solve the infinite-horizon discounted reward MDP problem
and was later extended to the average reward case in \cite{schwartz1993reinforcement}.
The main idea is to iteratively update an estimated action-value function until it converges to $Q^{*, \lambda}$.
The update rule in time step $t$ is given below
\begin{align}
    Q_{t+1}&(s, \alpha) \gets (1-\beta(t)) Q_t(s, \alpha) \notag \\
    &+ \beta(t) \left( C^{\lambda}(s, \alpha) - v_t + \min_{\alpha\in\mathcal{A}} Q_t(s', \alpha) \right), \label{eq:update_Q} \\
    v_{t+1} &\gets (1 - \beta(t)) v_t \notag \\
    &+ \beta(t) \left( C^{\lambda}(s, \alpha) + \min_{\alpha\in\mathcal{A}} Q_t(s', \alpha) - \min_{\alpha\in\mathcal{A}} Q_t(s, \alpha) \right), \label{eq:update_v}
\end{align}
where $Q_t(s, \alpha)$ and $v_t$ are the estimates of $Q^{*, \lambda}$ and $v^{*, \lambda}$ at slot $t$,
$\beta(t)$ is the learning rate at slot $t$,
and $s'$ is the observed state after we execute the greedy action $\alpha^*$ that minimizes $Q_t(s, \alpha)$.
It should be noted that if we adopt the greedy action $\alpha^*$ in every step,
the algorithm may stay in the local optimum we have already found.
To avoid this problem, the Q-learning algorithm takes random actions with a certain probability
and thus achieves a trade-off between exploration and exploitation.

While Q-learning ultimately results in the optimal control policy, its learning process is inefficient because
it learns the system's dynamics from scratch and does not harness existing knowledge of the system's dynamics.
Consequently, Q-learning usually suffers a long convergence time \cite{jaksch2010near, even2004learning}.
In our particular case, however, the reward function and part of state transition are known in advance.
Such prior knowledge allows us to enhance the algorithm's learning efficiency
with the help of PDS, which will be elaborated in the subsequent subsection.

\subsection{Post-Decision State Learning} \label{subsection:pds_learning}
Conventional Q-learning is designed for MDPs with totally unknown cost functions and transition probabilities.
In our problem, however, the system dynamics are partially known
(e.g. we know whether the HOL task will be completed once we choose an action)
so we only need to learn the unknown part.
To exploit the known information, we define the PDS as the intermediate system state after the action $\alpha(t)$ is conducted,
but before the random events happen.
Let the PDS at slot $t$ denoted by $\tilde{s}(t) = (\tilde{d}^r_i(t), \tilde{a}_i(t), \tilde{q}_i(t), \tilde{h}_i(t))_{i\in\mathcal{N}}$.
Since the channel gain is not affected by actions, we have $\tilde{h}_i(t) = h_i(t)$.
The queue length $\tilde{q}_i(t)$ is the number of queueing tasks after local processing and computation offloading,
so we have $\tilde{q}_i(t) = q_i(t) - b_i(t)$.
Similarly, $\tilde{d}^r_i(t)$ and $\tilde{a}_i(t)$ are the unfinished data in the HOL task and the AoI of WD $i$ after $\alpha(t)$.

With the introduction of PDS, the state transition process is divided into known and unknown parts.
The transition probabilities for the two parts are denoted by $p_k(\tilde{s} | s, \alpha)$ and $p_u(s'|\tilde{s}, \alpha)$, respectively.
Then the original transition probabilities are factored as
\begin{equation}
    p(s' | s, \alpha) = \sum_{\tilde{s}} p_u(s'|\tilde{s}, \alpha) p_k(\tilde{s}|s, \alpha).
\end{equation}
In our problem, the known dynamics are deterministic so $p_k(\tilde{s} | s, \alpha)$ degenerates to a function $\tilde{s} = f_k(s, \alpha)$.
The expression of $f_k(s, \alpha)$ is described in the previous paragraph.
Additionally, the unknown dynamics are independent to the action $\alpha$ so we have $p_u(s' | \tilde{s}, \alpha) = p_u(s' | \tilde{s})$.
Clearly, we only need to learn the unknown part of the transition process.

Similar to the optimal state-value function $V^{*, \lambda}$ in Q-learning,
we can also define the optimal PDS value function $\tilde{V}^{*, \lambda}(\tilde{s})$.
According to the relationship between $s$ and $\tilde{s}$, $\tilde{V}^{*, \lambda}(\tilde{s})$ and $V^{*, \lambda}(s)$ can be 
converted into each other:
\begin{gather}
    \tilde{V}^{*, \lambda}(\tilde{s}) = \sum_{s'\in\mathcal{S}} p_u(s'|\tilde{s}) V^{*, \lambda}(s'), \\
    V^{*, \lambda}(s) + v^{*, \lambda} = \min_{\alpha\in\mathcal{A}} \left\{ C^{\lambda}(s, \alpha) + \tilde{V}^{*, \lambda}(f_k(s, \alpha)) \right\}. \label{eq:pds_value_function}
\end{gather}
Substituting \eqref{eq:V_Q} into \eqref{eq:pds_value_function} results in
\begin{equation}
    Q^{*,\lambda}(s,\alpha) = C^{\lambda}(s, \alpha) + \tilde{V}^{*, \lambda}(f_k(s, \alpha)) - v^{*,\lambda}.
    \label{eq:pds_Q}
\end{equation}
By combining \eqref{eq:pds_Q} with \eqref{eq:optimal_policy} and using the fact that $v^{*,\lambda}$ is independent of $\alpha$,
we have
\begin{equation}
    {\pi}^{*,\lambda}(s) = \argmin_{\alpha\in\mathcal{A}} \left\{ C^{\lambda}(s, \alpha) + \tilde{V}^{*, \lambda}(f_k(s, \alpha)) \right\}.
\end{equation}
Hence, we can obtain the optimal policy when the optimal PDS value function is known.

The intuition behind PDS learning is very similar to that of Q-learning.
The main difference is we will use a sample average to approximate the optimal PDS value function $\tilde{V}^{*, \lambda}$ instead of $Q^{*, \lambda}$.
The major improvement is we exclude the known dynamics from $\tilde{V}^{*, \lambda}$ so we only learn the unknown information and thus enhance the learning efficiency.
Our algorithm is summarized as follows:
\begin{enumerate}
    \item At the beginning slot (i.e. $t = 0$), initialize the PDS value function $\tilde{V}^{0, \lambda}(\tilde{s}) \leftarrow 0$, the estimated optimal cost 
        $v^{0, \lambda} \leftarrow 0$, and the Lagrange multiplier $\lambda_i(0) \leftarrow 0$.
    \item At slot $t$, choose the greedy action
        \begin{equation}
            \alpha(t) = \argmin_{\alpha\in\mathcal{A}} \left\{ C^{\lambda}(s(t), \alpha) + \tilde{V}^{t, \lambda}( f_k(s(t), \alpha) ) \right\}.
            \label{prob:greedy}
        \end{equation}
        The algorithm that finds the greedy action is described in our previous work \cite{he2022age}.
    \item Observe the random events at slot $t$ and obtain $s(t+1)$.
    \item Compute the value of state for $s=s(t)$ and $s=s(t+1)$
        \begin{align}
            V^{t, \lambda}(s) = \min_{\alpha\in\mathcal{A}} \Big\{ C^{\lambda}(s, \alpha)
            + \tilde{V}^{t, \lambda}(f_k(s, \alpha)) \Big\} - v^{t, \lambda}.
            \label{eq:V_V_tilde}
        \end{align}
    \item Update the PDS value function and the estimated optimal cost
        \begin{align}
            &\tilde{V}^{t+1, \lambda}(\tilde{s}(t)) \leftarrow (1 - \beta(t)) \tilde{V}^{t, \lambda}(\tilde{s}(t)) \notag \\
            &\qquad + \beta(t) V^{t, \lambda}(s(t+1)) \label{eq:PDS_update_V} \\
            &v^{t+1, \lambda} \leftarrow (1 - \beta(t)) v^{t, \lambda} + \beta(t) \times \bigl( C^{\lambda}(s(t), \alpha(t)) \notag \\
            &\qquad + V^{t, \lambda}(s(t+1)) - V^{t, \lambda}(s(t)) \bigr) \label{eq:PDS_update_v}
        \end{align}
        where $\tilde{s}(t)$ is the PDS at slot $t$.
    \item Update the Lagrange multiplier 
        \begin{equation}
            \lambda_i(t+1) \leftarrow \Lambda \left[ \lambda_i(t) + \eta(t) (E_i(t) - E^{max}_i) \right], \label{eq:update_lambda}
        \end{equation}
        where $\Lambda$ is an operator that projects $\lambda$ onto $[0, \lambda^{max}]$ for some large enough $\lambda^{max} > 0$
        and $\eta(t)$ is a time-varying learning rate.
    \item Update the time index (i.e. $t \leftarrow t+1$) and go to the second step.
\end{enumerate}

In conventional Q-learning, random actions are necessary to explore the unknown cost and transition probability functions that depend on the action.
In our problem, however, the unknown dynamics are independent of the action (i.e. $p_u(s'|\tilde{s}, \alpha) = p_u(s'|\tilde{s})$)
so we do not need exploration in PDS learning.

Another difference from the conventional Q-learning is our algorithm learns the PDS value function $\tilde{V}^{*,\lambda}(\tilde{s}(t))$
instead of the action-value function $Q^{*,\lambda}(s, \alpha)$.
Since $\tilde{V}^{*,\lambda}(\tilde{s}(t))$ only depends on the PDS state, it has much fewer
possible inputs than $Q^{*,\lambda}(s,\alpha)$.
As a result, the learning process of $\tilde{V}^{*,\lambda}(\tilde{s}(t))$ converges much more quickly
than that of $Q^{*,\lambda}(s,\alpha)$.
The reason that we only need to learn $\tilde{V}^{*,\lambda}(\tilde{s}(t))$ is the reward function $C^{\lambda}(s,\alpha)$
and the transition function $f_k(s,\alpha)$ are known in our problem, so having $\tilde{V}^{*,\lambda}(\tilde{s}(t))$ is sufficient to
calculate the greedy action, as demonstrated in \eqref{prob:greedy}.

The following theorem shows that the PDS learning converges to the optimal policy when the learning rates satisfy some mild conditions.
\begin{theorem}
    The estimated PDS value function $\tilde{V}^{t, \lambda}$ converges to $\tilde{V}^{*, \lambda}$ in stationary environments
    when the sequence of learning rates $\beta(t)$ satisfy
    \begin{equation}
        \sum_{t=0}^{\infty} \beta(t) = \infty \mbox{ and } \sum_{t=0}^{\infty} \beta(t)^2 < \infty.
        \label{eq:rate_condition}
    \end{equation}
    Additionally, if $\eta(t)$ has the same properties \eqref{eq:rate_condition} as $\beta(t)$,
    then the Lagrange multiplier $\lambda(t)$ also converges to $\lambda^*$ when the following additional requirements are satisfied
    \begin{equation}
        \sum_{t=0}^{\infty} \beta(t)^2 + \eta(t)^2 < \infty \mbox{ and } \lim_{t\to\infty} \frac{\eta(t)}{\beta(t)} \to 0.
    \end{equation}
    \label{theorem:pds_convergence}
\end{theorem}
\begin{IEEEproof}
    The proof follows the same approach as in the proof of Theorem 2 in \cite{mastronarde2012joint} and is omitted for brevity.
\end{IEEEproof}

In the above theorem, the convergence of the PDS learning requires a stationary environment.
In practice, however, if we keep $\beta(t)$ and $\eta(t)$ away from zero,
the estimated PDS value function and optimal cost are continuously updated based on the new experience
and thus enable our algorithm to track non-stationary Markovian dynamics.

\subsection{Existing Problems of Tabular PDS Learning} \label{subsection:problems}
The PDS learning algorithm proposed in the previous subsection is based on the tabular Q-learning.
It requires the state and action spaces to be sufficiently small so that approximate value functions can be represented 
in a tabular form, such as arrays or tables.
Generally, tabular reinforcement learning algorithms inherently exhibit the following drawbacks:
\begin{itemize}
    \item \emph{Cannot handle continuous state and action spaces.}
        To let the approximate value functions be represented by arrays or tables, the state and action spaces must have finite elements.
        Therefore, the tabular reinforcement learning methods cannot handle problems with continuous state or action spaces.
        Although we can discretize the state and action spaces, it necessarily leads to suboptimal solutions.
    \item \emph{Cannot handle high-dimensional problems.}
        In MDP problems,
        the size of state and action spaces grows exponentially with respect to the dimensions of system states and control actions.
        For high-dimensional problems, the number of possible states and actions is extremely large hence the convergence time of the learning process
        is prohibitively long.
    \item \emph{Cannot handle unseen states.}
        In practice, we may encounter unseen states when applying reinforcement learning algorithms.
        For example, in our problem, a new channel state $h_i(t)$ may be observed if the wireless environment changes.
        In this case, the value of $\tilde{V}^{t, \lambda}( f_k(s(t), \alpha) )$ in \eqref{prob:greedy} is unknown
        so we are unable to obtain the greedy action.
\end{itemize}



To address the above issues,
recent works \cite{mnih2013playing, lillicrap2015continuous, mnih2016asynchronous} have tried to combine
DNNs with reinforcement learning and achieved great success in a diverse set of applications.
This new class of algorithms is often referred to as Deep Reinforcement Learning (DRL).
Different from tabular reinforcement learning, DRL uses DNNs to represent value functions.
Since DNNs are able to approximate continuous functions and process high-dimensional data \cite{sarker2021deep}, 
DRL algorithms can be used to solve problems with continuous and high-dimensional state and action spaces.
Moreover, the generalization ability of DNNs enables DRL algorithms to 
choose appropriate actions even for unseen states.
Due to the above advantages, we will extend our tabular PDS learning to a DRL algorithm in the next section.

\section{DDPG-Based Deep PDS Learning} \label{section:dpds}
As the first well-known DRL algorithm, Deep Q-Network (DQN) \cite{mnih2013playing} combines tabular Q-learning
with DNNs and is widely employed to tackle various control problems.
However, DQN is not applicable to our problem because it can only handle discrete action spaces while the decision variables
in our problem take continuous values.
To circumvent this issue, we develop our DRL algorithm based on the Deep Deterministic Policy Gradient (DDPG)~\cite{lillicrap2015continuous} method, which excels in handling continuous action spaces.

Nonetheless, integrating DDPG with PDS learning poses non-trivial challenges.
Firstly, DDPG (as well as its variants) is primarily designed for MDPs with \textit{discounted} rewards, whereas our problem revolves around \textit{average} rewards.
Recent research \cite{xu2021optimal, naik2019discounted} has demonstrated considerable difficulty in adapting algorithms initially designed for discounted reward MDPs to average reward MDPs. Consequently, we have to tailor the original DDPG algorithm to align with the average reward setting.
Secondly, DDPG employs a critic network to estimate the optimal action-value function $Q^{*,\lambda}$. 
In contrast, our PDS learning only maintains the PDS value function, thus necessitating a redesign of the critic network's architecture and update rules.
Thirdly, the success of DRL algorithms usually requires substantial adjustment and fine-tuning.
Therefore, we need to make several modifications to the standard DDPG framework to adapt it to our specific problem.

To tackle these aforementioned challenges, in this section, we first propose the modified DDPG for average reward MDPs. 
Following this, we introduce our novel DDPG-based deep PDS learning algorithm. 
At the end of this section, we present various implementation details of the algorithm.

\subsection{Average-Reward DDPG} \label{subsection:a_ddpg}
DDPG is a well-known DRL algorithm based on deterministic policy gradient and DQN. To deal with continuous state and action spaces, DDPG utilizes 
two DNNs to estimate the optimal policy function 
$\pi^{*,\lambda}(s)$ and the optimal action-value function $Q^{*,\lambda}(s,\alpha)$. These DNNs are typically referred to as the primary actor network (whose parameter is denoted by $\theta^{\pi}$) and the primary critic network (whose parameter is denoted by $\theta^Q$).
Then the two estimate functions can be represented by $\pi^{\lambda}(s\, | \, \theta^{\pi})$ and 
$Q^{\lambda}(s,\alpha\, | \, \theta^Q)$, respectively.
For average reward MDPs, we also need to estimate $v^{*,\lambda}$, and the estimated value is denoted by $v^{\lambda}$.
The update of $Q^{\lambda}(s,\alpha)$ and $v^{\lambda}$ follows the same rule as in \eqref{eq:update_Q} and \eqref{eq:update_v}.
The main difference is, instead of directly modifying the tabular data, DDPG updates the estimated action-value function
by optimizing the DNN's parameter $\theta^Q$.
To this end, we define the loss of the primary critic network as
\begin{equation}
    L^{\lambda}(\theta^Q) = \mathbb{E} \left[ \left( Q^{\lambda}(s,\alpha \, | \, \theta^Q) - y \right)^2 \right]
\end{equation}
where
\begin{equation}
    y = C^{\lambda}(s,\alpha) - v^{\lambda} + Q^{\lambda}(s', \pi^{\lambda}(s' \, | \, \theta^{\pi}) \, | \, \theta^Q).
    \label{eq:y}
\end{equation}
Then the parameter $\theta^Q$ can be optimized by minimizing $L^{\lambda}(\theta^Q)$ with various optimizers like Stochastic Gradient Descent (SGD).
As for the primary actor network, it has been proved in \cite{silver2014deterministic} that the gradient of $\theta^{\pi}$
can be computed as 
\begin{equation}
    \nabla_{\theta^{\pi}}J \approx \mathbb{E} \left[ \nabla_{\alpha} Q^{\lambda}(s, \alpha \, | \, \theta^Q)|_{\alpha=\pi^{\lambda}(s)}
    \nabla_{\theta^{\pi}} \pi^{\lambda}(s \, | \, \theta^{\pi}) \right],
    \label{eq:actor_update}
\end{equation}
where $J$ is the expected long-term reward.

To improve the algorithm's performance, DDPG utilizes two extra techniques during the training process.
Firstly, DDPG stores historical experiences in a replay buffer and uniformly samples a minibatch for training at each step.
This ensures the training data are decorrelated and can greatly stabilize the training process.
Secondly, when minimizing the critic loss $L^{\lambda}(\theta^Q)$, we try to optimize $\theta^Q$ so that 
$Q^{\lambda}(s,\alpha \, | \, \theta^Q)$ is as close to $y$ as possible.
However, according to \eqref{eq:y}, the value of $y$ depends on the same parameter $\theta^Q$ we are trying to optimize
and this may cause stability problems during the update.
To avoid this issue, DDPG introduces a copy of the primary actor and critic network, denoted by
$\pi_T^{\lambda}(s \, | \, \theta^{\pi}_T)$ and $Q_T^{\lambda}(s, \alpha \, | \, \theta^{Q}_T)$.
In the literature, $\pi_T^{\lambda}$ and $Q_T^{\lambda}$ are usually referred to as the target actor and critic network.
The parameters of $\pi_T^{\lambda}$ and $Q_T^{\lambda}$ are updated to slowly track the primary networks:
\begin{align}
    \theta^{Q}_T &= \omega \theta^Q + (1-\omega) \theta_T^{Q} \label{eq:target_critic} \\
    \theta^{\pi}_T &= \omega \theta^{\pi} + (1-\omega) \theta_T^{\pi}, \label{eq:target_actor}
\end{align}
where $\omega \ll 1$ is a constant that determines the tracking speed.
For average reward MDPs, we also need to set a target value for $v^{\lambda}$, and the corresponding update formula is:
\begin{equation}
    {v}_T^{\lambda} = \omega v^{\lambda} + (1-\omega) {v}_T^{\lambda}. \label{eq:target_average_reward}
\end{equation}
With the help of the above target values, $y$ can be computed as
\begin{equation}
    y = C^{\lambda}(s,\alpha) - {v}_T^{\lambda} + {Q}_T^{\lambda}(s', {\pi}_T^{\lambda}(s' \, | \, \theta_T^{\pi}) \, | \, \theta_T^{Q}),
\end{equation}
which no longer depends on $\theta^Q$.

\subsection{DDPG-based Deep PDS Learning}
\begin{figure*}[!t]
    \centering
    \includegraphics[width=0.7\textwidth]{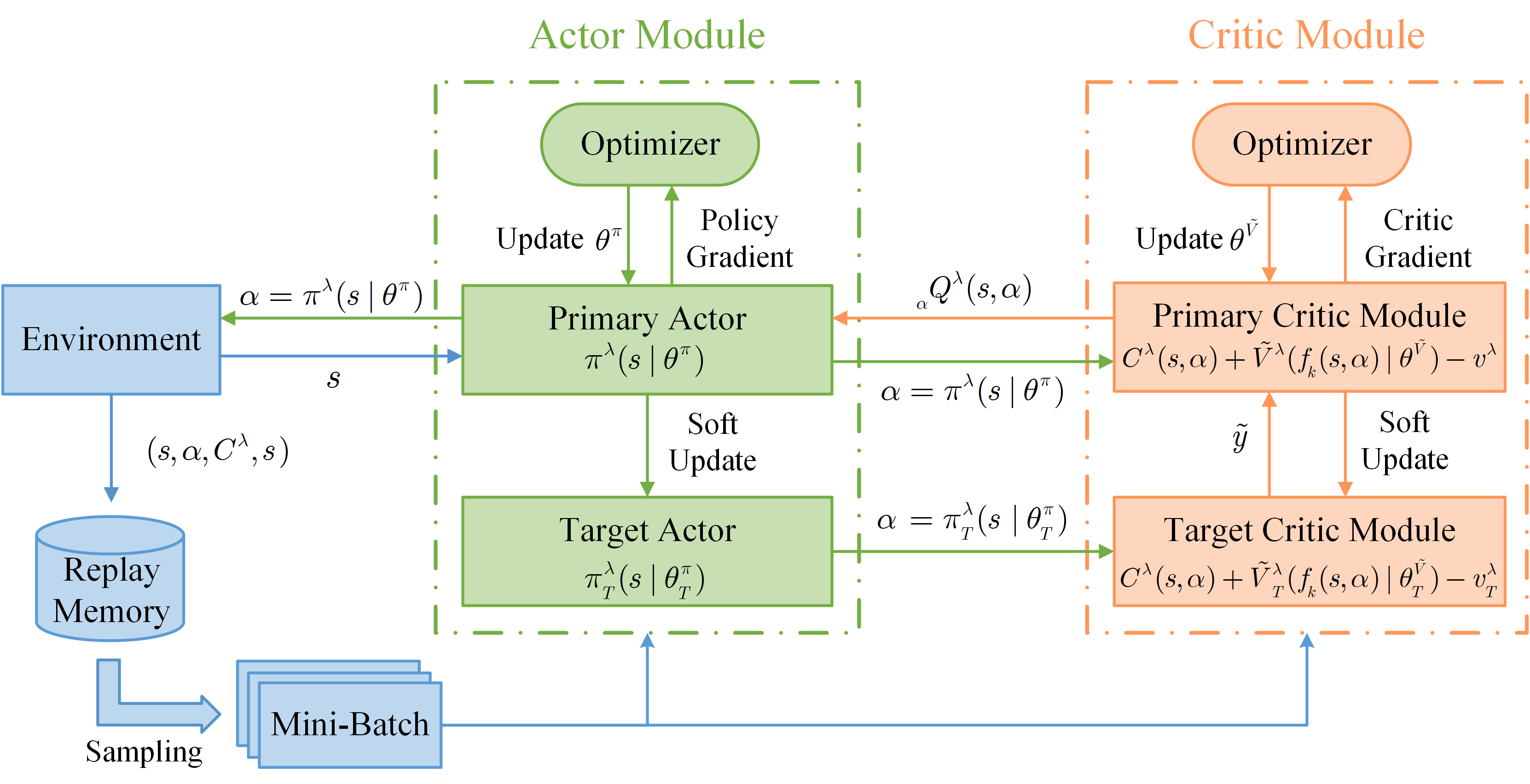}
    \caption{Algorithm structure of DDPG-based deep PDS learning.}
    \label{fig:ddpg}
\end{figure*}
Similar to the Q-learning algorithm, DDPG is also designed for MDPs with completely unknown system dynamics. To enhance the learning efficiency, we combine DDPG with the PDS learning approach introduced in Section \ref{subsection:pds_learning}. The structure of our algorithm is depicted in Fig. \ref{fig:ddpg}. The main difference from DDPG is we use a DNN to estimate the optimal PDS value function $\tilde{V}^{*,\lambda}(\tilde{s})$
instead of the optimal action-value function $Q^{*,\lambda}(s,\alpha)$. The DNN is referred to as the primary value network (whose parameter is denoted by $\theta^{\tilde{V}}$), and it is accompanied by a target value network (whose parameter is denoted by $\theta_T^{\tilde{V}}$).
Then the estimate of $\tilde{V}^{*,\lambda}(\tilde{s})$ can be represented by $\tilde{V}^{\lambda}(\tilde{s} \, | \, \theta^{\tilde{V}})$ and its target value is $\tilde{V}_T^{\lambda}(\tilde{s} \, | \, \theta_T^{\tilde{V}})$. Accordingly, the update of the target value network follows 
\begin{equation}
    \theta^{\tilde{V}}_T = \omega \theta^{\tilde{V}} + (1-\omega) \theta^{\tilde{V}}_T.
    \label{eq:target_value}
\end{equation}

As illustrated in \eqref{eq:pds_Q},
$\tilde{V}^{\lambda}(\tilde{s} \, | \, \theta^{\tilde{V}})$, $C^{\lambda}(s,\alpha)$, 
$f_k(s,\alpha)$, and $v^{\lambda}$ together constitute an estimate of $Q^{*,\lambda}(s,\alpha)$.
For convenience, we call them the primary critic module.
To optimize $\theta^{\tilde{V}}$, we need to define a new loss for the primary value network.
Based on the PDS update rule \eqref{eq:PDS_update_V}, the expression of $V^{t,\lambda}(s)$ \eqref{eq:V_V_tilde},
and the fact that our greedy action is produced by the actor network, the loss of the primary value network
is given by:
\begin{equation}
    \tilde{L}^{\lambda}(\theta^{\tilde{V}}) = \mathbb{E} \left[ \left( \tilde{V}^{\lambda}(\tilde{s} \, | \, \theta^{\tilde{V}})
    - \tilde{y} \right)^2\right],
    \label{eq:new_loss}
\end{equation}
where
\begin{equation*}
    \tilde{y} = C^{\lambda}(s', \pi^{\lambda}_T(s' \, | \, \theta^{\pi}_T)) 
    + \tilde{V}^{\lambda}_T(f_k(s', \pi^{\lambda}_T(s' \, | \, \theta^{\pi}_T)) \, | \, \theta^{\tilde{V}}_T) - v^{\lambda}_T.
\end{equation*}
Notice that we have used target values when calculating the loss.
For the primary actor network, we can still use \eqref{eq:actor_update} to compute the gradient.
However, since we no longer have the primary critic network,
the gradient $\nabla_{\alpha} Q^{\lambda}(s, \alpha \, | \, \theta^Q)|_{\alpha=\pi^{\lambda}(s)}$ in \eqref{eq:actor_update}
should be replaced by 
\begin{equation}
    \nabla_{\alpha} ( C^{\lambda}(s, \alpha) + \tilde{V}^{\lambda}(f_k(s,\alpha) \, | \, \theta^{\tilde{V}}) )|_{\alpha=\pi^{\lambda}(s)}.
    \label{eq:new_gradient}
\end{equation}

Based on the previous discussion, the DDPG-based deep PDS learning algorithm is summarized in Algorithm \ref{alg:deep_pds}.
Our algorithm commences by initializing the parameters of the primary value network and the primary actor network (i.e. $\theta^{\tilde{V}}$ and $\theta^{\pi}$), respectively. 
Additionally, the parameters of the corresponding target networks are initialized as $\theta^{\tilde{V}}_T = \theta^{\tilde{V}}$ and $\theta^{\pi}_T = \theta^{\pi}$. 
The replay buffer is emptied, and values of the average reward and its target, the Lagrange multipliers, and the maximum training steps are also set at this point.
After the initialization, the algorithm enters the training phase, which lasts for $T_{max}$ steps. 
In each step $t$, the primary actor network is employed to obtain the greedy action $\alpha(t)$ given the current state $s(t)$. 
After executing $\alpha(t)$, we observe the next state and store the new experience tuple $(s(t), \alpha(t), C^{\lambda}(s(t), \alpha(t)), s(t+1))$ into the replay buffer.
If the replay buffer is full, the oldest experience tuple is discarded.
When the replay buffer has sufficient experience tuples, we randomly sample a minibatch to train the primary networks.
At last, we update the Lagrange multipliers $\lambda_i$ to ensure the time average energy consumption constraint is satisfied.

\begin{algorithm}[t]
    \caption{DDPG-based Deep PDS Learning} \label{alg:deep_pds}
    \begin{algorithmic}[1]
        \State \textbf{Initialize:}
        Primary networks parameters $\theta^{\tilde{V}}$ and $\theta^{\pi}$, target networks parameters
        $\theta^{\tilde{V}}_T \leftarrow \theta^{\tilde{V}}$ and $\theta^{\pi}_T \leftarrow \theta^{\pi}$,
        average reward $v^{\lambda}$ and its target $v^{\lambda}_T \leftarrow v^{\lambda}$,
        replay buffer $\gets \varnothing$, Lagrange multiplier $\lambda_i$, maximum training steps $T_{max}$.
        \For{$t \gets 1, T_{max}$}
        \State \textbf{Action Selection:} $\alpha(t) = \pi^{\lambda}(s(t) \, | \, \theta^{\pi})$.
        \State \parbox[t]{\dimexpr\linewidth-\algorithmicindent}{
            \textbf{Observation:} Execute action $\alpha(t)$, observe the random events and obtain $s(t+1)$.
        \strut}
        \State \parbox[t]{\dimexpr\linewidth-\algorithmicindent}{
            \textbf{Update Replay Buffer:} Store the new experience $(s(t), \alpha(t), 
            C^{\lambda}(s(t), \alpha(t)), s(t+1))$ in the buffer.
        \strut}
        \State \parbox[t]{\dimexpr\linewidth-\algorithmicindent}{
            \textbf{Sampling:} Randomly sample a minibatch of tuples from the replay buffer.
        \strut}
        \State \parbox[t]{\dimexpr\linewidth-\algorithmicindent}{
            \textbf{Update Critic:} Calculate loss for the sampled tuples according to \eqref{eq:new_loss}
            and update the primary critic network by minimizing $\tilde{L}^{\lambda}(\theta^{\tilde{V}})$.
        \strut}
        \State \parbox[t]{\dimexpr\linewidth-\algorithmicindent}{
            \textbf{Update Actor:} Update the primary actor network according to \eqref{eq:actor_update}
            and \eqref{eq:new_gradient}.
        \strut}
        \State \parbox[t]{\dimexpr\linewidth-\algorithmicindent}{
            \textbf{Update Average Reward:} Update the average reward according to \eqref{eq:PDS_update_v}.
        \strut}
        \State \parbox[t]{\dimexpr\linewidth-\algorithmicindent}{
            \textbf{Update Targets:} Update target values with \eqref{eq:target_value}, \eqref{eq:target_actor},
            and \eqref{eq:target_average_reward}.
        \strut}
        \State \parbox[t]{\dimexpr\linewidth-\algorithmicindent}{
            \textbf{Update Lagrange Multiplier:} Update $\lambda_i$ with \eqref{eq:update_lambda}.
        \strut}
        \EndFor
    \end{algorithmic}
\end{algorithm}

\subsection{Implementation Details} \label{subsection:details}
While DRL algorithms hold the promise of ultimately converging to the optimal results, achieving this convergence often demands substantial adjustments and fine-tuning, such as setting appropriate hyperparameters, designing suitable reward functions, managing high-dimensional state and action spaces, and choosing proper neural network architectures.
In this subsection, we will introduce several techniques that we have employed in the implementation of the DDPG-based deep PDS learning algorithm.

\subsubsection{Redesign of Cost Function} \label{subsubsection:redesign_cost}
The cost function $C^{\lambda}(s,\alpha)$ defined in \eqref{eq:cost_function} has two terms, corresponding to AoI and energy consumption, respectively.
The trade-off between these two terms is determined by $\lambda_i$.
When $\lambda_i$ is too small, the algorithm focuses solely on reducing AoI, which usually leads to a violation of energy constraints.
In contrast, when $\lambda_i$ is too large, the AoI may grow to a very large value.
To achieve a good balance, we must update $\lambda_i$ dynamically to approach the optimal $\lambda^*_i$.
However, due to the heterogeneity of WDs, it is usually very difficult to find a proper initial value and learning rate for $\lambda_i$.
Moreover, changing the value of $\lambda_i$ will alter the distribution of data stored in the replay buffer,
hence makes the training process of our algorithm unstable.
To address this issue, we replace the second term in $C^{\lambda}(s,\alpha)$ with $\sum_{i\in\mathcal{N}} \lambda_i \max(0, E_i(s, \alpha) - E^{max}_i)$.
The main change is we will not yield a positive reward when the energy consumption falls below the energy budget.
Therefore, even when $\lambda_i$ takes a relatively large value, 
our algorithm will maintain energy consumption around the energy budget instead of continually reducing it.
Hence, our algorithm can work well as long as $\lambda_i$ is sufficiently large, which
means we no longer need to obtain an accurate estimate of $\lambda^*_i$.
Since the estimation accuracy is not crucial, we can also reduce the learning rate of $\lambda_i$.
This ensures the data distribution in the replay buffer remains relatively stable, 
thereby improving training stability.

Another problem of the original cost function is that the current AoI $a_i(s,\alpha)$ is determined by the current state $s$
and is independent of the action $\alpha$.
When we calculate the gradient of the actor network using \eqref{eq:new_gradient},
we have $\nabla_{\alpha} C^{\lambda}(s,\alpha) = \sum_{i\in\mathcal{N}} \lambda_i \nabla_{\alpha} \max(0, E_i(s,\alpha) - E^{max}_i)$.
Therefore, the first term in \eqref{eq:new_gradient} only instructs the actor to reduce the cost of energy consumption.
This also means the actor learns how to minimize AoI solely through the second term
$\nabla_{\alpha} \tilde{V}^{\lambda}(f_k(s,\alpha) \, | \, \theta^{\tilde{V}})$.
However, the value $\tilde{V}^{\lambda}(f_k(s,\alpha) \, | \, \theta^{\tilde{V}})$ is estimated by another DNN, i.e. the value network.
This gives rise to two problems.
Firstly, the value network may not provide accurate estimates in the early stages, leading to training instability.
Secondly, when computing gradients using backpropagation, passing through an additional DNN can lead to gradient vanishing issues \cite{hochreiter1998vanishing}, which makes $\nabla_{\alpha} \tilde{V}^{\lambda}(f_k(s,\alpha) \, | \, \theta^{\tilde{V}})$
smaller than $\nabla_{\alpha} C^{\lambda}(s,\alpha)$.
Consequently, the update of the actor network will lean toward prioritizing energy consumption reduction.
To solve the two problems, we replace the current AoI $a_i(s, \alpha)$ in $C^{\lambda}(s,\alpha)$
with the \emph{expected AoI after action}.
Specifically, let $p^a_i$ and $\bar{d}_i$ be the average task arrival rate and average data size of tasks at WD $i$
obtained through historical experiences.
Apparently, processing a task will reduce WD $i$'s AoI by $1/p^a_i$ in expectation.
Hence, the expected reduction of AoI after the action $\alpha$ is $a^-_i(s, \alpha) = \frac{d^l_i(t)+d^o_i(t)}{\bar{d}_i} \frac{1}{p^a_i}$.
Then the expected AoI after action is $a_i(s,\alpha) - a^-_i(s,\alpha)$ and our cost function becomes
\begin{align*}
  C^{\lambda}(s, \alpha) = \sum_{i\in\mathcal{N}} &\left( a_i(s,\alpha) - a^-_i(s,\alpha) \right) \\
                         &+ \sum_{i\in\mathcal{N}} \lambda_i \max\left(0, E_i(s,\alpha) - E^{max}_i \right).
\end{align*}

\subsubsection{Normalization of Decision Variables}
In our problem, the decision variables have significantly different ranges of values.
For example, the CPU frequency $f_i(t)$ may take several GHz while the transmission power $P_i(t)$ is usually less than $1$W.
Such differences can potentially hinder the convergence of the DNN training process \cite{sola1997importance}.
To achieve better results, we normalize decision variables to the $[0,1]$ interval.
For example, instead of $f_i(t)$, our actor network will output $\hat{f}_i(t) \in [0,1]$, then
$f_i(t)$ can be calculated as $\hat{f}_i(t) \times f^{max}_i$.
However, we cannot directly use this method to obtain $W_i(t)$, because $W_i(t)$ also needs to satisfy constraint \eqref{amp:bandwidth}.
To address this issue, we first observe that increasing bandwidth will only result in a better objective value, 
so the constraint \eqref{amp:bandwidth} must be tight at the optimal solution.
Hence, we can replace constraint \eqref{amp:bandwidth} with $\sum_{i\in\mathcal{N}} W_i(t) = W^{max}$ for all $t\in\mathcal{T}$.
By feeding all $\hat{W}_i(t)$ to a softmax function we have $\sum_{i\in\mathcal{N}} softmax(\hat{W}_1, \dots, \hat{W}_N)_i = 1$.
Then we can obtain a feasible $W_i(t)$ by multiplying $W^{max}$ with the outputs of the softmax function.

\subsubsection{Recalculate Cost During Training}
During the training process of DDPG, we use past experience stored in the replay buffer to update DNNs.
The underlying assumption is that these historical data follow the same distribution. 
However, in our problem, the value of $\lambda$ evolves continuously, leading to different costs for the same state-action pair.
Hence, the assumption that the data distribution remains unchanged no longer holds.
If we use costs calculated based on outdated $\lambda$ to update the current DNNs, it may result in incorrect update directions.
Fortunately, our cost function is known.
Therefore, during the training process, we can recalculate the cost in the experience tuple based on the current $\lambda$,
thus enhancing training stability.
This also means we no longer to record the observed costs in the historical experiences.

\subsubsection{Neural Network Architecture}
\begin{figure}[t]
\centering
\subfloat[Actor Network]{\includegraphics[width=0.15\textwidth]{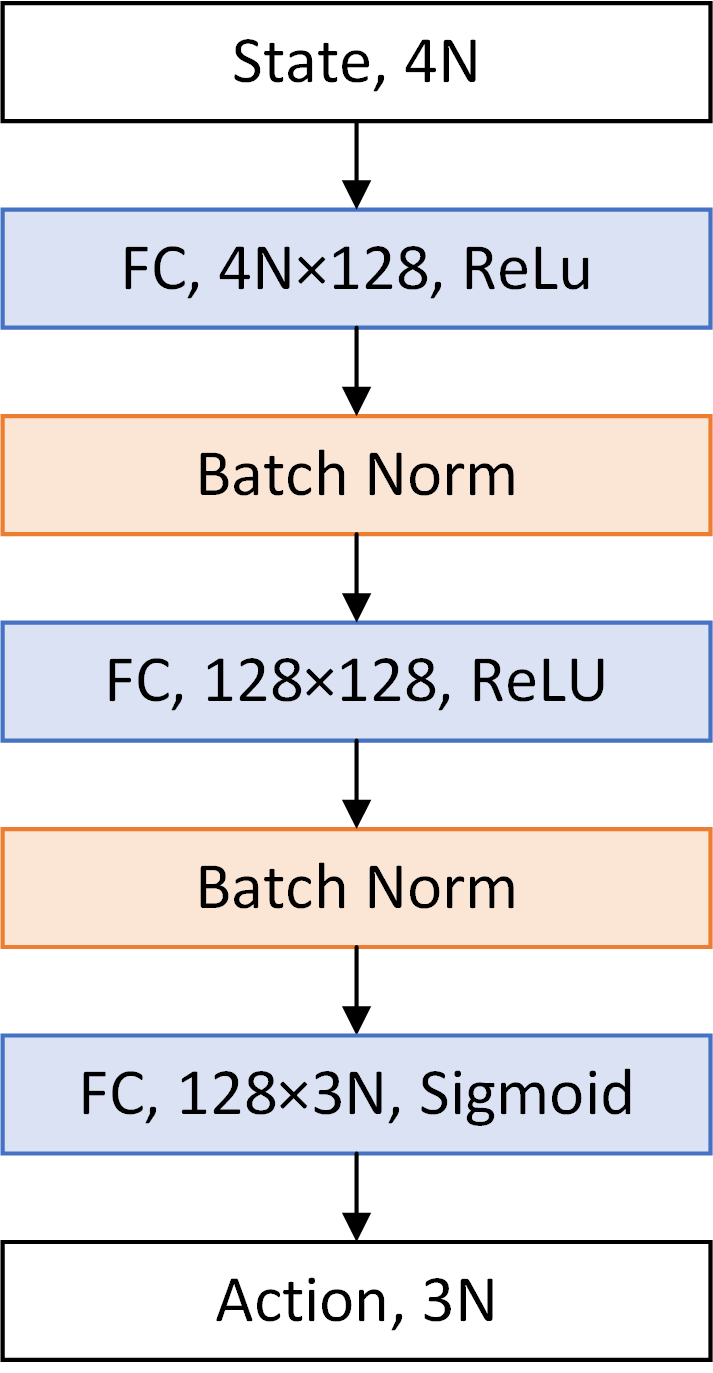} \label{fig:actor}}
\hfil
\subfloat[Value Network]{\includegraphics[width=0.15\textwidth]{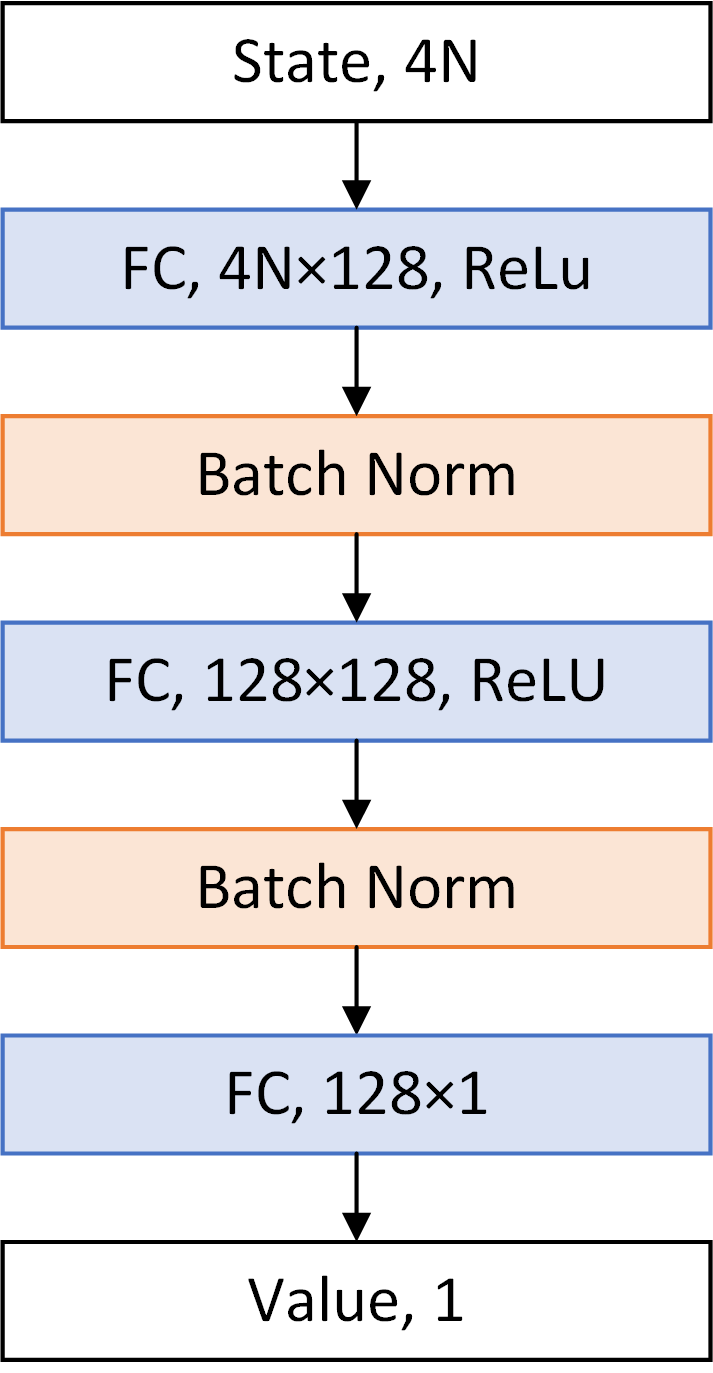} \label{fig:value}}
\caption{Neural network architecture used in our algorithm.}
\label{fig:nna}
\end{figure}

The neural network architecture employed in our algorithm is presented in Fig. \ref{fig:nna}. 
Taking the actor network as an example.
The input to this network is the system's state, which is a collection of the individual state of each WD. 
Given that the state of each WD is represented as a four-tuple, the dimension of the system's state is $4N$. 
This system state is then fed into a fully-connected (FC) layer comprising 128 neurons, each incorporating a rectified linear unit (ReLU) activation function. 
The outputs from these neurons undergo normalization via a batch norm layer to enhance the training stability and convergence of DNNs.
Subsequently, the normalized results are fed into yet another FC layer and batch norm layer with the same configuration.
The final layer of the actor network is an FC layer with $3N$ neurons where
Sigmoid activation functions are utilized to ensure that the outputs fall within the interval $[0,1]$.

The architecture of the value network is vary similar to that of the actor network.
The only difference is we only use one neuron (without activation function) at the final layer.
To train these two DNNs, we employ two Adam optimizers with learning rate $0.001$ (for actor network) and $0.002$ (for value network),
respectively.

\subsubsection{Computational Complexity}
In this subsection, we analyze the computational complexity of the proposed DPDS algorithm.
We specifically focus on the computational workload arising from the forward and backward passes of DNNs, 
as the computational demands of other operations, such as sampling training batches and updating target values, are deemed negligible.

The computational complexity of DNNs is typically quantified by the number of Floating Point OPerations (FLOPs). 
Since the FLOPs associated with activation functions and batch norm layers are relatively minor compared to those of Fully Connected (FC) layers, 
our analysis exclusively considers the latter. 
For a single forward pass, the FLOPs of an FC layer with $n_i$ inputs and $n_o$ outputs amount to $2 n_i n_o$. 
Accordingly, the FLOPs for the actor network and the value network per forward pass are expressed as 
$FLOP_a = 2 \times 4N \times 128 + 2 \times 128 \times 128 + 2 \times 128 \times 3N = 1792N + 32768$ and 
$FLOP_v = 2 \times 4N \times 128 + 2 \times 128 \times 128 + 2 \times 128 = 1024N + 33024$, respectively. 
According to the results in \cite{sevilla2022estimating}, the FLOPs associated with the backward pass are approximately twice those of the forward pass.

For DRL algorithms, the computational complexity during the execution and training phases is different. 
When deploying DPDS in realistic systems, obtaining the scheduling decisions in each iteration necessitates only a single forward pass of the actor network. 
Even for a wireless network with $N = 100$ WDs, the corresponding workload is around $2.1 \times 10^5$ FLOPs. 
Considering that modern GPUs, such as the Nvidia H100, can deliver up to $10^{15}$ FLOPs per second \cite{nvidia2022h100}, 
the computational workload due to action selection is very light.

The training process requires slightly more computational power. 
Specifically, when updating the value network, we first need to calculate the loss according to equation \eqref{eq:new_loss}, 
necessitating a forward pass of the target actor network, the target value network, and the value network. 
The obtained loss is then employed to update the value network through the backward pass. 
Hence the total FLOPs involved in a single update of the value network is around $FLOP_a + FLOP_v + FLOP_v + 2\times FLOP_v = 5888N+164864$. 
The FLOPs for updating the actor network can be derived similarly. 
In total, the computational workload for the training process in a single iteration is approximately $12288N + 296192$ FLOPs.

\section{Numerical Results} \label{section:simulation}
\begin{figure}[t]
    \centering
    \includegraphics[width=0.35\textwidth]{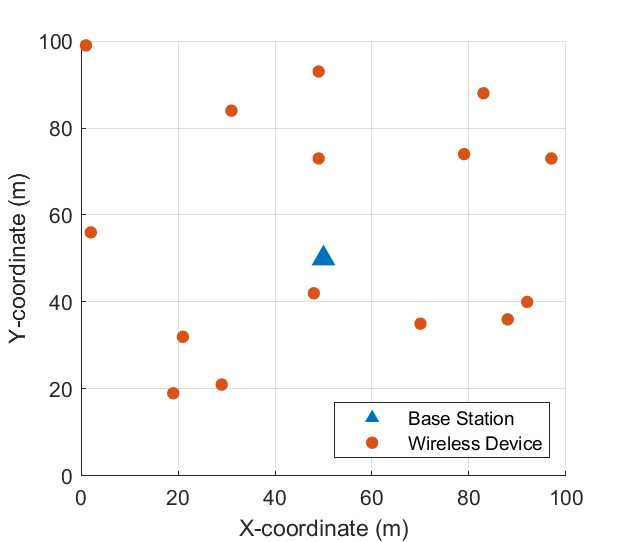}
    \caption{Locations of WDs and the BS.}
    \label{fig:location}
\end{figure}
In this section, numerical simulations are conducted to evaluate the performances of our proposed algorithm.
We consider a MEC network where $N=15$ WDs are randomly located in a $100$m$\times 100$m square
and the BS is in the center of the square.
The locations of WDs and the BS are shown in Fig. \ref{fig:location}.
Each WD is equipped with a processor with a maximum frequency of $f^{max}_i = 2$GHz.
The generation of new tasks follows a Poisson process with an arrival rate of $0.3$.
The data size of each task is uniformly distributed in the range of $[20\mbox{Kb}, 50\mbox{Kb}]$.
The processing of each bit of data demands $\kappa = 1\mbox{K}$ CPU cycles,
and the energy efficiency of WD's processor is $\gamma = 10^{-28}$.
The channel gain between WD $i$ and the BS follows the Rayleigh fading model,
which is given by $h_i(t) = 10^{-3} d^{-\epsilon}_i \rho(t)$ \cite{wu2019online, ju2013throughput},
where $d_i$ is the distance between WD $i$ and the BS,
$\epsilon$ is the pathloss exponent,
and $\rho(t)$ is an exponentially distributed random variable with a unit mean which represents the short-term fading.
In our simulation, we assume $\epsilon = 3.8$.
The noise power at the BS is $\sigma^2 = 10^{-11}$.
Each WD has a maximum transmission power of $P^{max}_i = 1\mbox{W}$ and all WDs share $W^{max} = 20$MHz wireless bandwidth.
We require the long-term energy consumption of each WD cannot exceed $E^{max}_i = 0.1$W.
The utilized learning rates are $\beta(t) = 1/\sqrt{t}$ and $\eta(t) = 10^2 \times \min(1, 10/\log(t+1))$.
The duration of each slot is $10$ms and each simulation lasts for $10^5$ slots.
For clarity, the AoI and energy consumption data presented in this section are averaged over all WDs.

For comparison purpose, we also attempted to implement the tabular PDS learning algorithm proposed in Section \ref{subsection:pds_learning}.
Unfortunately, the tabular PDS learning algorithm failed to learn any effective policy for a fairly long time and the outputs seem to be totally random. 
The main reason for this problem is the space of the PDS is prohibitively large. 
Recall that the PDS for an individual WD is a four-tuple: $(\tilde{d}^r_i, \tilde{a}_i, \tilde{q}_i, \tilde{h}_i)$. 
For simplicity, we assume the max values of $\tilde{a}_i$ and $\tilde{q}_i$ are $50$ and $20$, respectively. 
Moreover, we discretize the data size $\tilde{d}^r_i$ and channel gain $\tilde{h}_i$ into $50$ and $20$ different levels. 
Then the number of possible PDS for an individual WD is $50 \times 50 \times 20 \times 20 = 10^6$. 
By utilizing the cost function decomposition technique proposed in Section \uppercase\expandafter{\romannumeral 4}-F of \cite{he2022age},
the number of possible PDS of the entire system is linear (instead of exponential) with respect to the number of WDs. 
Therefore, the overall number of possible PDS is approximately $15\times 10^6=1.5\times 10^7$. 
However, this number is still too large for the tabular PDS learning algorithm to converge.
The above observation highlights the limitations of tabular PDS learning when handling practical situations with large state spaces,
whcih also serves as a validation of our analysis in Section \ref{subsection:problems}.

\subsection{Runtime Performance of Deep PDS Learning}
\begin{figure}[t]
\centering
\subfloat[AoI]{\includegraphics[width=0.38\textwidth]{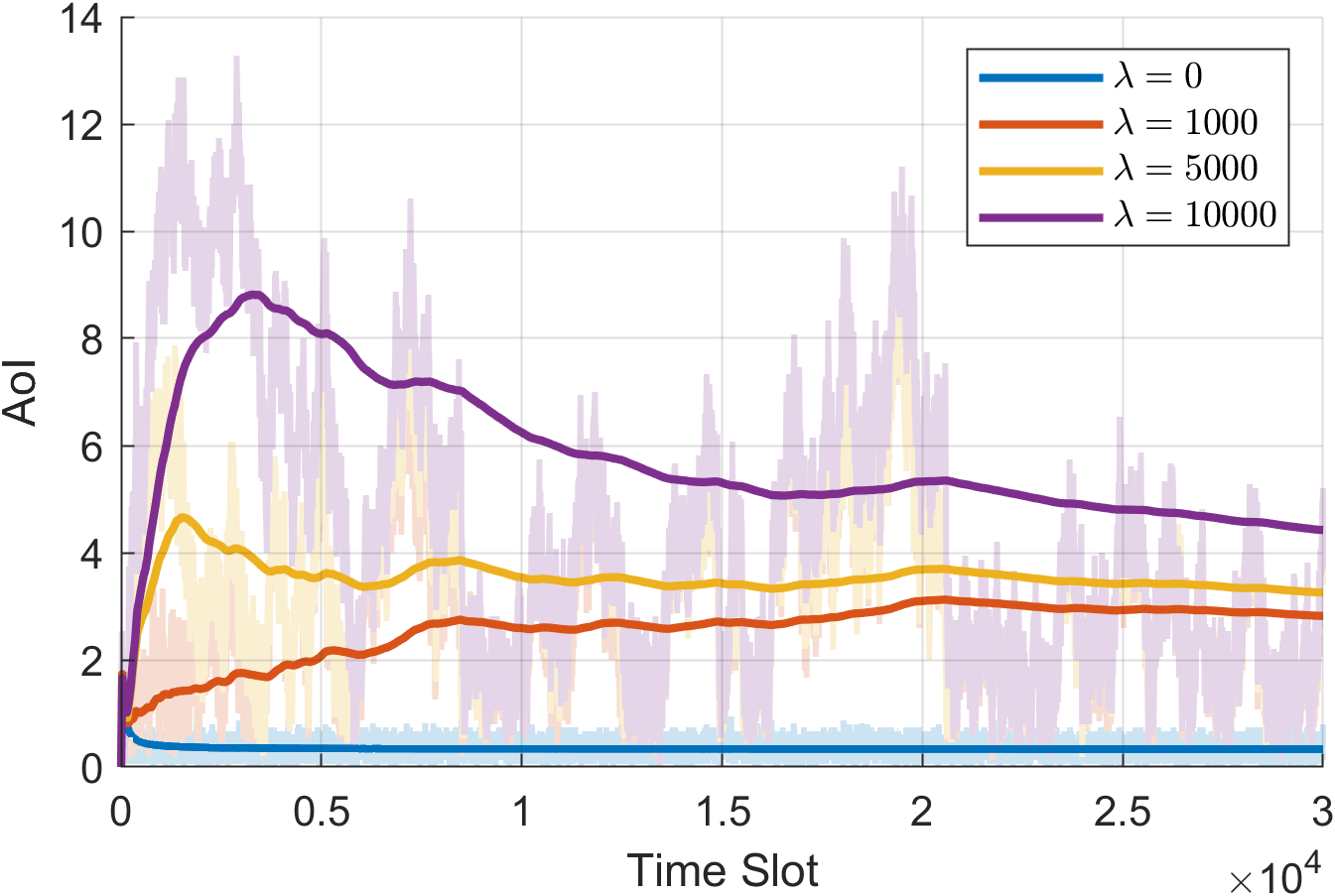} \label{time_dynamics_vep_A}}
\vfil
\subfloat[Energy Consumption]{\includegraphics[width=0.38\textwidth]{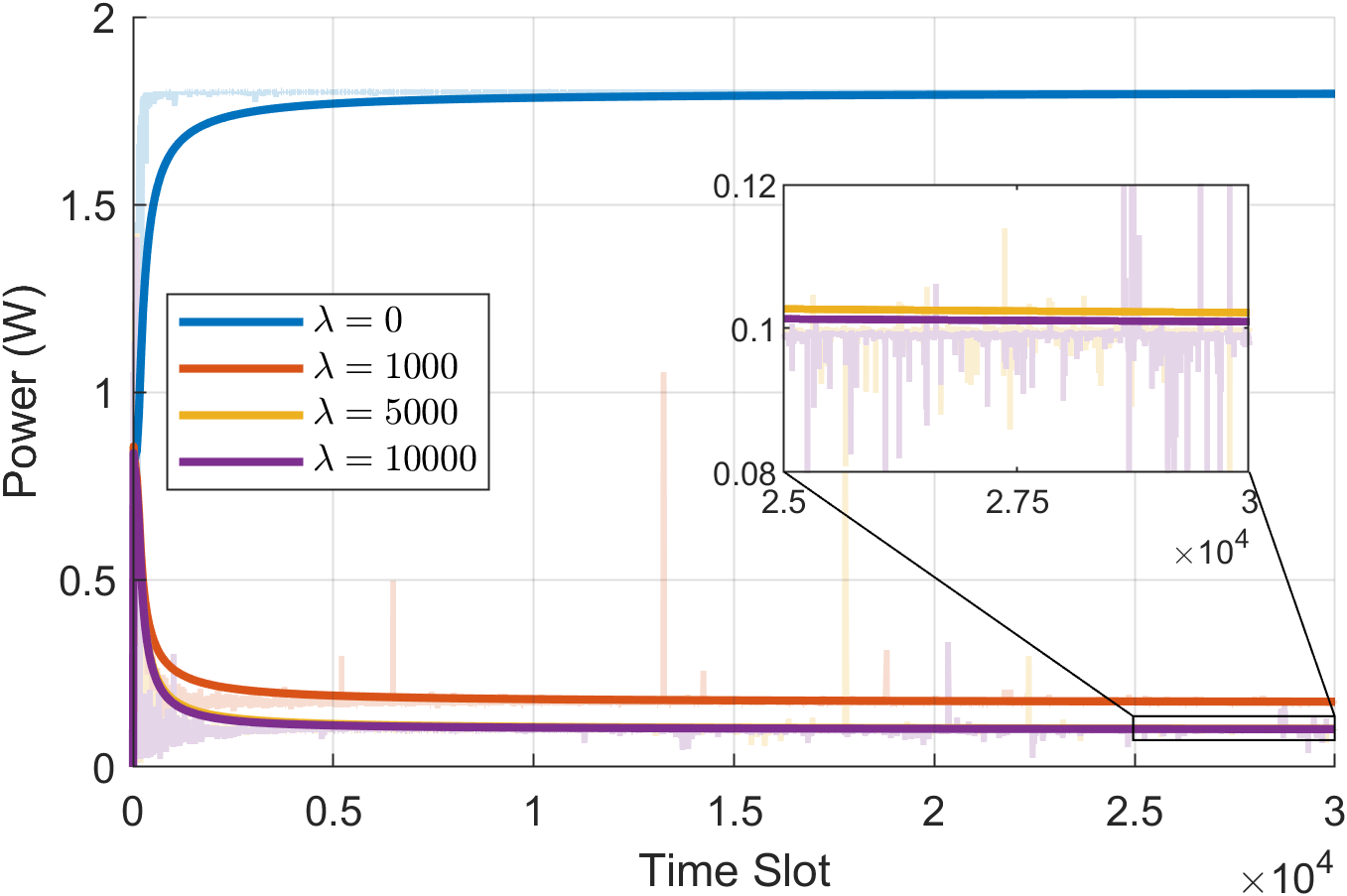} \label{time_dynamics_vep_E}}
\caption{Runtime performance under different initial values of $\lambda$.}
\label{time_dynamics_lam}
\end{figure}

The AoI (measured in slots) and energy consumption under different initial values of $\lambda$ are presented in Fig. \ref{time_dynamics_lam},
where the semi-transparent lines represent real-time data, while the solid lines represent their temporal averages.
Due to the small changes in the latter half, we have only displayed the data for the first $3\times 10^4$ slots.
Clearly, as the value of $\lambda$ increases, the energy consumption is reduced, but at the cost of higher AoI.
When the initial value of $\lambda$ is set to either 0 or 1000, our algorithm falls short of meeting the energy constraint. 
However, when the initial value of $\lambda$ is sufficiently large (e.g., $5000$ or $10000$), the energy consumption can be maintained around the budget,
which aligns with our previous analysis in Section \ref{subsubsection:redesign_cost}.
While it is true that the energy consumption remains almost identical whether the initial value of $\lambda$ is set to 5000 or 10000, 
the former yields a notably smaller AoI. Consequently, for all subsequent experiments, we choose to set the initial value of $\lambda$ to 5000.

Another observation from Fig. \ref{time_dynamics_lam} is that our algorithm appears to be highly sensitive to the initial value of $\lambda$.
For example, when $\lambda$ is initialized to $0$, the algorithm will focus on minimizing AoI at the beginning.
After that, the value of lambda increases rapidly as the energy consumption significantly exceeds the budget.
However, the algorithm seems to refuse to adjust its strategy accordingly and continues to maintain high energy consumption.
This indicates that the initial updates of the actor network are crucial
and subsequent adjustments are insufficient to reverse the strategy that has already been formed.

\subsection{In Comparison with Benchmark Algorithms}
\begin{figure}[t]
\centering
\subfloat[AoI]{\includegraphics[width=0.37\textwidth]{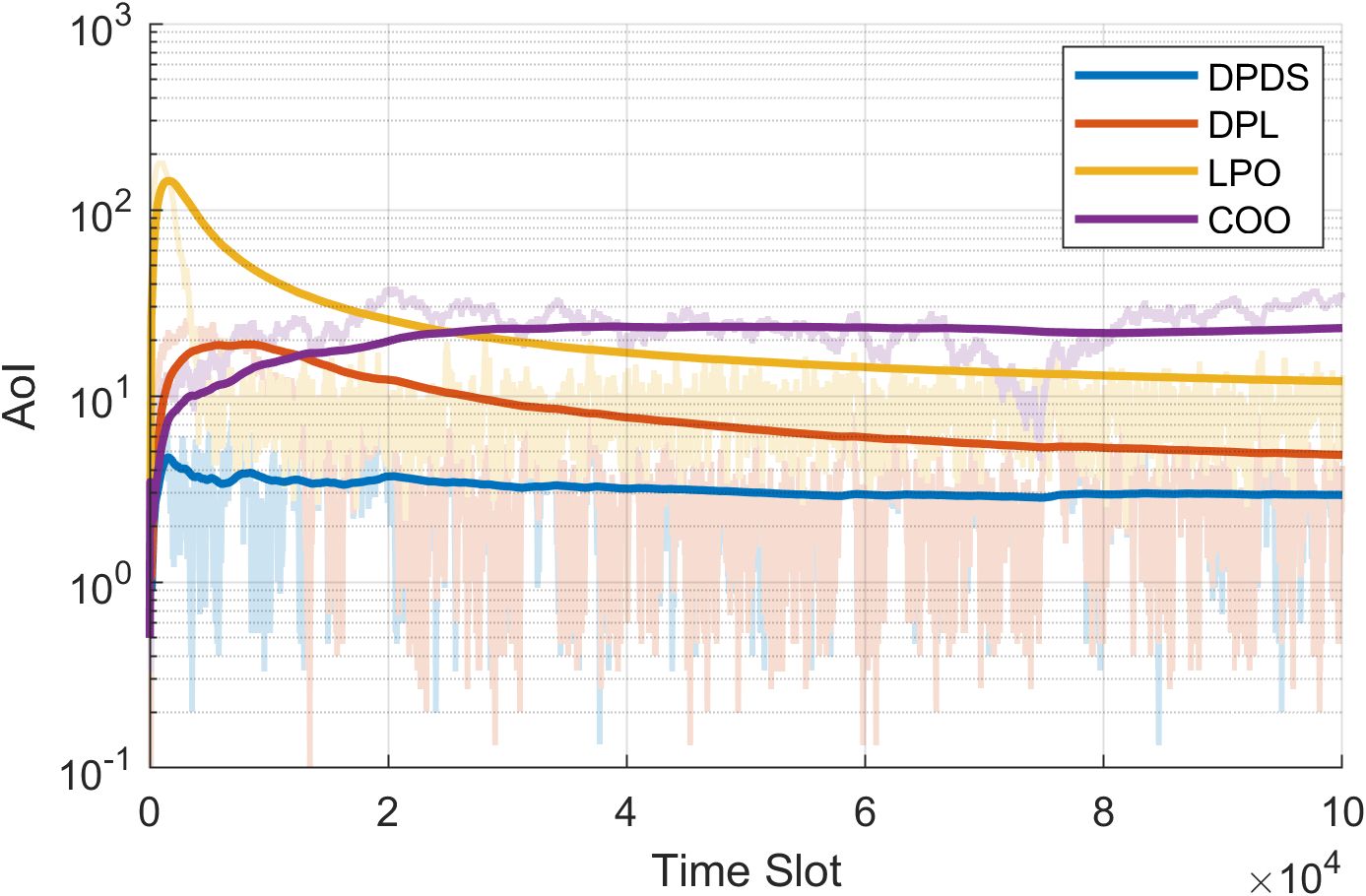} \label{time_dynamics_benchmark_A}}
\vfil
\subfloat[Energy Consumption]{\includegraphics[width=0.37\textwidth]{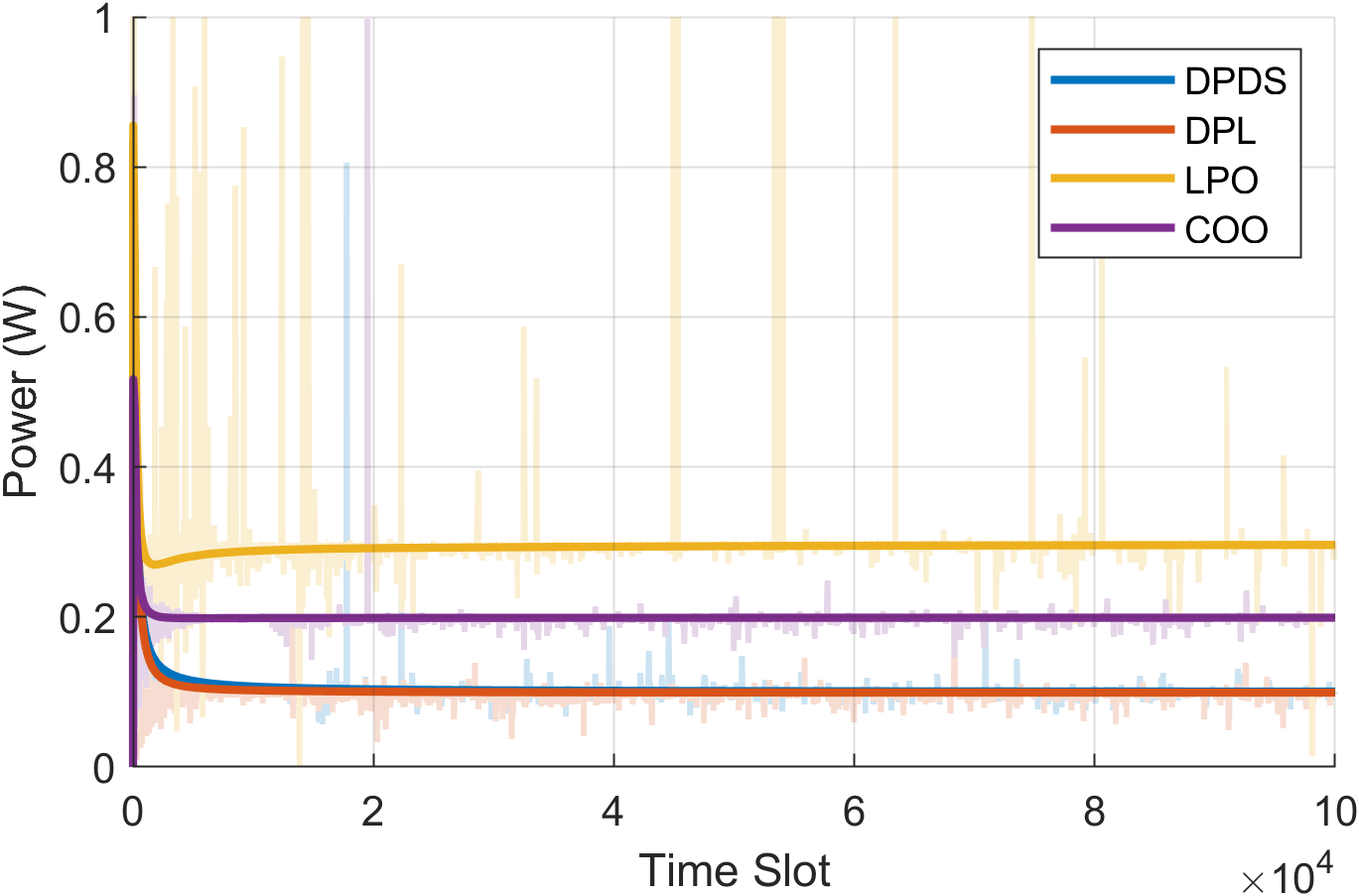} \label{time_dynamics_benchmark_E}}
\caption{Run-time performance of different algorithms.}
\label{time_dynamics_benchmark}
\end{figure}

In order to demonstrate the advantages of Deep PDS learning (DPDS), we compare it with the following benchmarks:
\begin{itemize}
    \item Local Processing Only (LPO): the computation data are always processed locally on WDs;
    \item Computation Offloading Only (COO): the computation tasks are fully offloaded to the BS;
    \item Delay-based PDS Learning (DPL): same as the proposed algorithm except that the objective is to minimize the total latency of tasks.
\end{itemize}
The selection of LPO and COO as benchmark algorithms is motivated by their frequent use in related literature studying computation offloading in MEC systems
\cite{song2019age, li2021age, jiang2023age, yang2020computation, zhao2019computation}. 
In particular, they serve as common baselines to highlight the advantages due to the strategic combination of local computing and computation offloading. 

The run-time performance of all algorithms is presented in Fig. \ref{time_dynamics_benchmark}.
Since COO and LPO only use a single computation method, their energy efficiency is relatively lower. 
To ensure the convergence of their AoI, we have increased their energy budgets by two times and three times, respectively.
Fig. \ref{time_dynamics_benchmark_E} indicates that the energy consumption of all algorithms can converge quickly to their budgets.

Fig. \ref{time_dynamics_benchmark_A} displays the variation in AoI for all algorithms,
where the logarithmic scale is used for the y-axis.
Even if we have significantly increased the energy budgets for LPO and COO,
their average AoI is still approximately four times and eight times that of DPDS,
which fully demonstrates the advantages brought by partial offloading.
Furthermore, even with partial offloading, the average AoI for DPL is significantly higher than that of DPDS, 
thus verifying our earlier argument that minimizing latency is not equivalent to optimizing information freshness.

\begin{figure}[t]
\centering
\subfloat[AoI]{\includegraphics[width=0.37\textwidth]{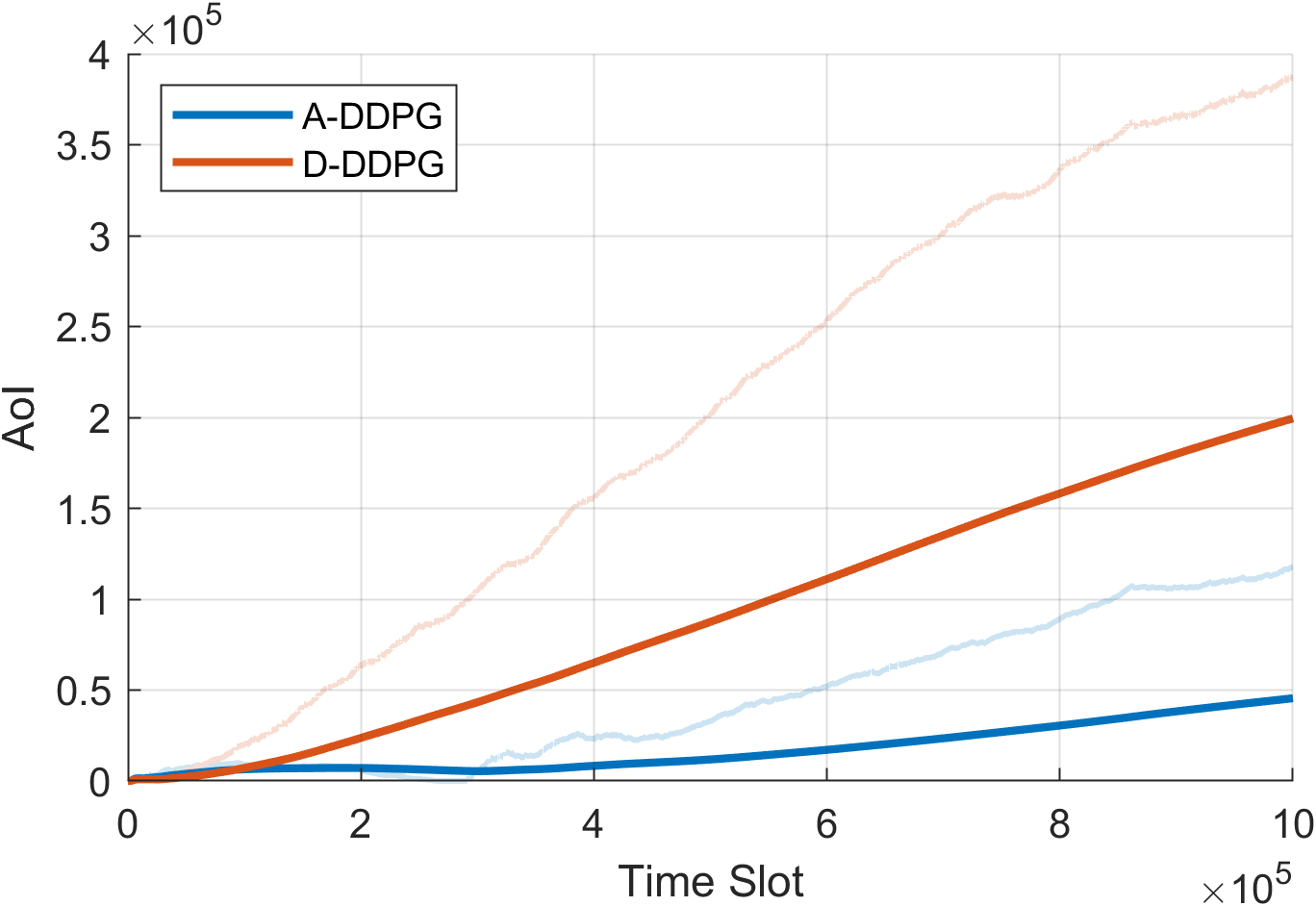} \label{ddpg_aoi}}
\vfil
\subfloat[Energy Consumption]{\includegraphics[width=0.37\textwidth]{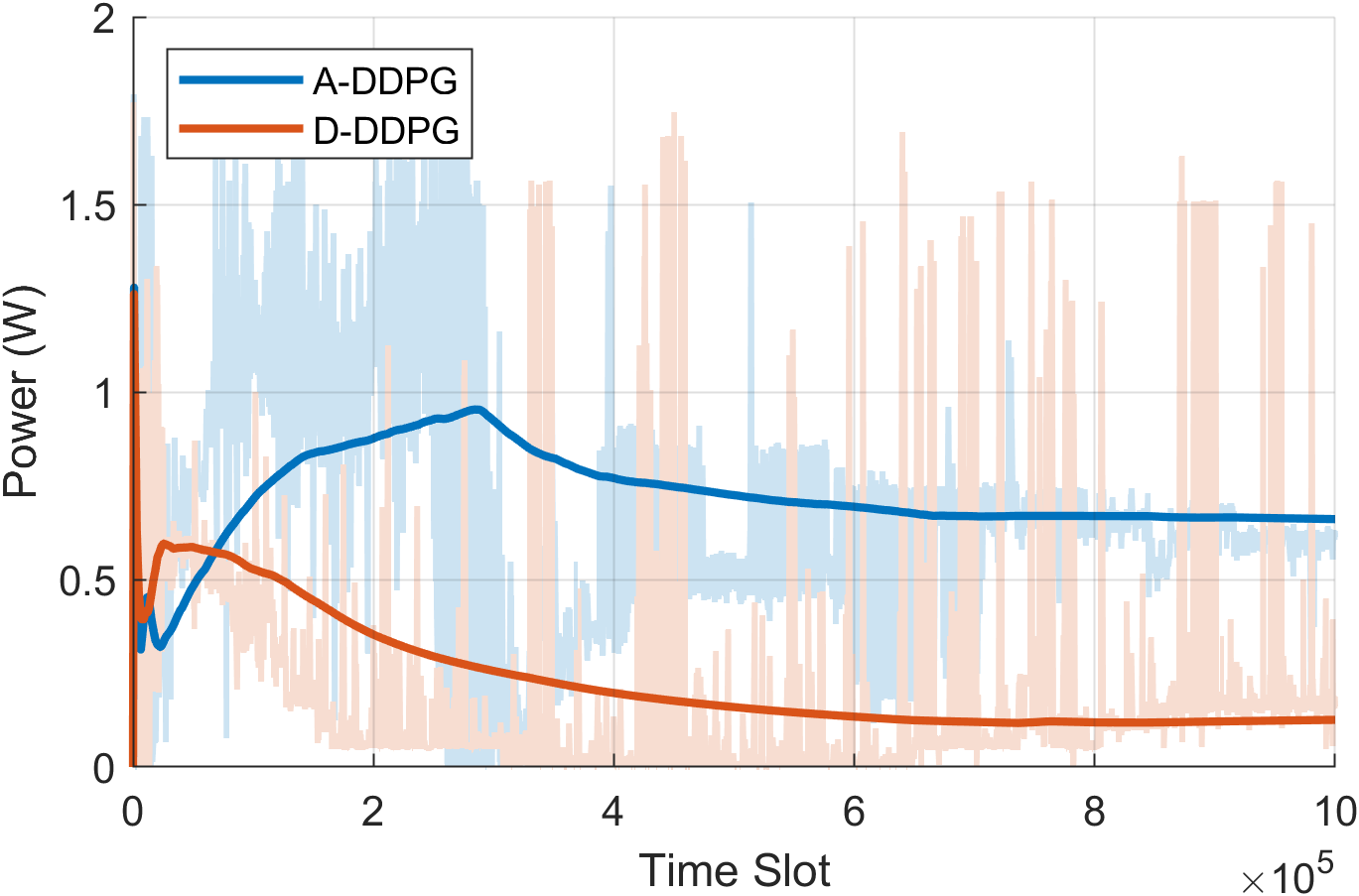} \label{ddpg_power}}
\caption{Run-time performance of different algorithms.}
\label{fig:sim_ddpg}
\end{figure}
The proposed DPDS algorithm can be regarded as a fusion of PDS and DDPG.
For comparison purpose, we implement the average-reward DDPG (A-DDPG) presented in Section \ref{subsection:a_ddpg}
and the conventional discounted-reward DDPG (D-DDPG).
Since the convergence speed of both algorithms is slow, we have extended the duration of the experiment to $10^6$ time slots.
However, even after such a long period, the algorithms failed to converge to an effective scheduling policy.
As shown in Fig. \ref{fig:sim_ddpg}, the AoI for both algorithms continues to increase to very large values.
Although A-DDPG's AoI increases relatively slowly, its average energy consumption is much higher than the budget.
Additionally, we have also observed that the two algorithms are particularly sensitive to neural architecture and parameter settings, 
making them prone to training instability.
The above observation fully demonstrates the advantages of DPDS over DDPG,
which is primarily attributed to the introduction of PDS and the techniques presented in Section \ref{subsection:details}.

\subsection{Performance under Different Numbers of WDs}
\begin{figure}[t]
  \centering
  \includegraphics[width=0.37\textwidth]{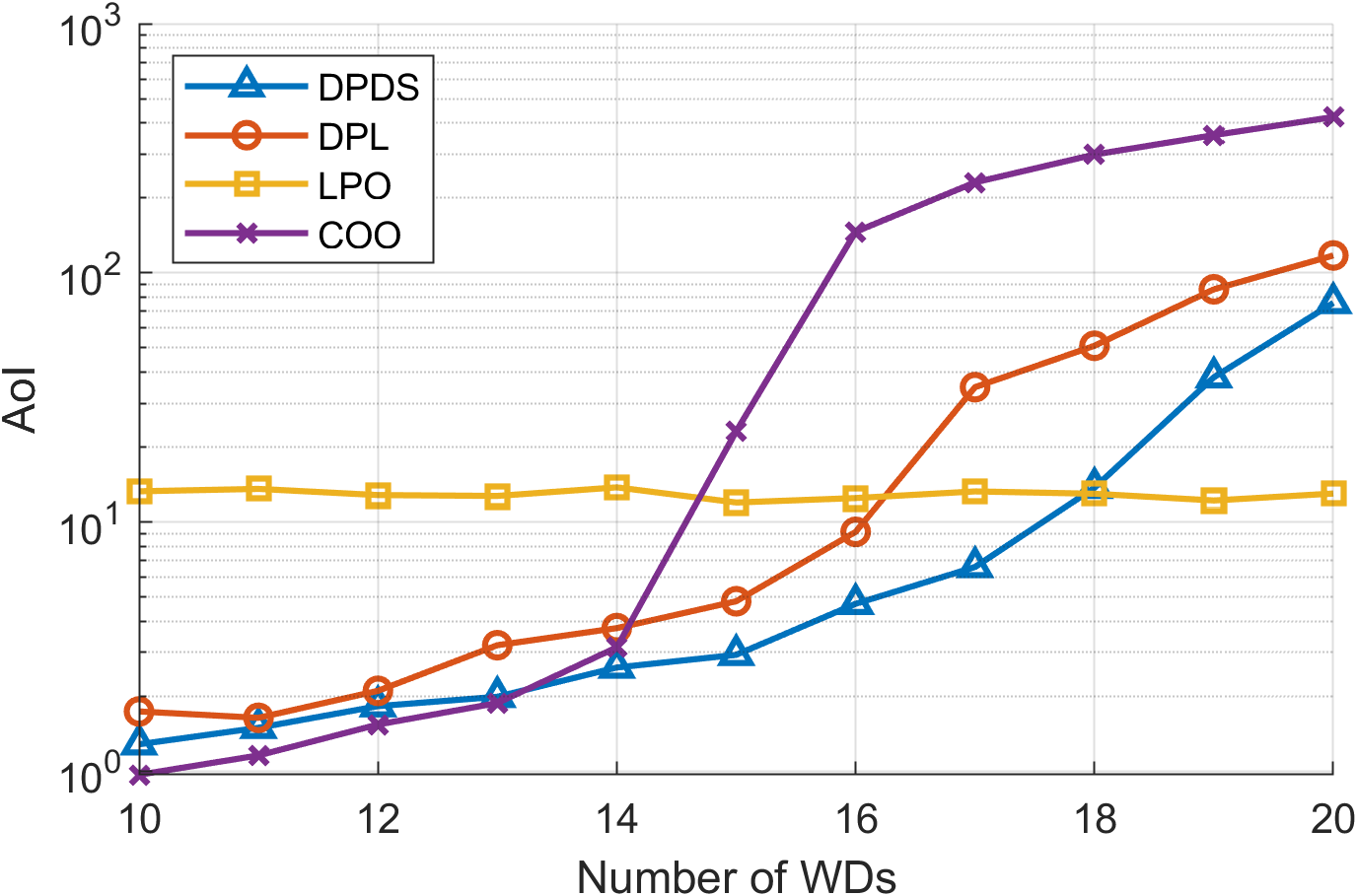}
  \caption{Average AoI under different numbers of WDs.}
  \label{fig:scale}
\end{figure}
Fig. \ref{fig:scale} shows the average AoI of different algorithms under different numbers of WDs.
Notice that the energy budgets of COO and LPO are still two times and three times that of DPDS.
The rest parameters remain unchanged as described at the beginning of this section.
In our problem, since the total spectrum bandwidth of the BS is fixed, having more WDs means that each WD can occupy fewer spectrum resources.
According to Shannon's formula, the wireless transmission rate is linear to the spectrum bandwidth.
Therefore, for scheduling schemes like COO, which can only perform computation offloading, the number of WDs has a significant impact on AoI.
In contrast, the average AoI of LPO remains the same regardless of the number of WDs.
For DPDS and DPL, since they partially rely on computation offloading, 
the increase in the number of WDs also leads to an increase in AoI, but the overall impact is much smaller than that of COO.

\subsection{Performance under Different Energy Budgets}
\begin{figure}[t]
  \centering
  \includegraphics[width=0.37\textwidth]{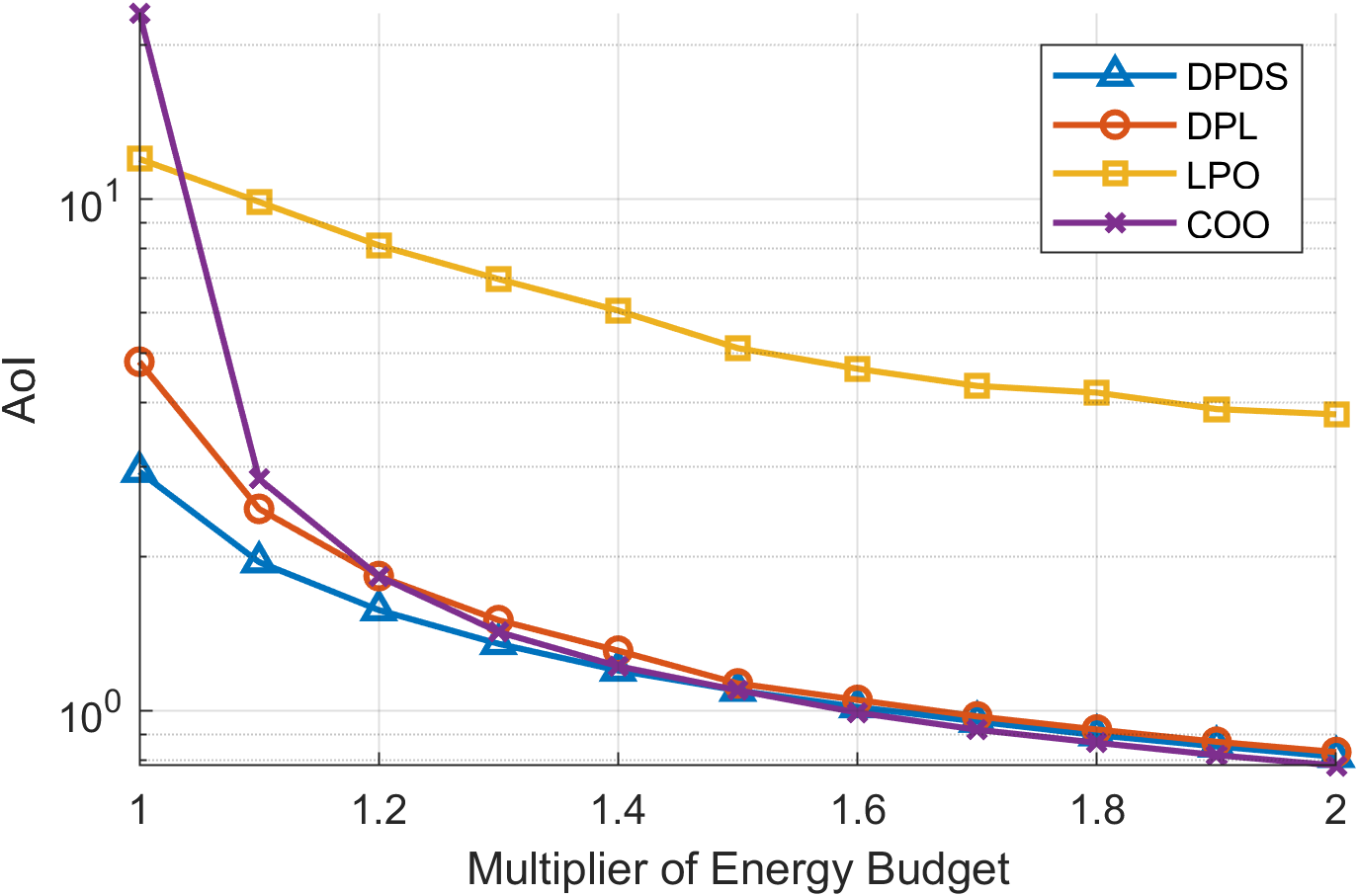}
  \caption{Average AoI under different numbers of WDs.}
  \label{fig:energy}
\end{figure}
To analyze the impact of the energy consumption constraint on AoI, we conduct experiments under different energy budgets.
Specifically, we gradually increase the energy budget of each algorithm with a multiplier.
Therefore, the energy budgets for COO and LPO are consistently two times and three times that of DPDS.
Our experiment results are displayed in Fig. \ref{fig:energy}.
As expected, the AoI for all algorithms decreases with the increase in the energy budget, but the rate of decrease continues to diminish.
Notice that both DPL and DPDS aim to process all queueing tasks as soon as possible. 
However, due to the difference in the definition of AoI and delay, the optimal scheduling policy of tasks for these two metrics is different. 
This explains why DPL produces sub-optimal AoI compared to DPDS. 
Nevertheless, as $E_i^{max}$ keeps growing, the processing capabilities of WDs become large enough so that almost all tasks can be completed immediately. 
In this case, the scheduling policy of both algorithms is very similar and results in nearly identical AoI.

Among the four algorithms, COO is the most sensitive to the energy budget, especially when the energy budget is relatively low. 
For example, when we initially increase the energy budget by $10\%$, the average AoI of COO decreases from $23.1$ to $2.8$, 
showing an improvement of almost an order of magnitude.
Combining with the data presented in the previous subsection, we can slightly increase the energy budget when the number of WDs increases,
thus offsetting the negative impact of insufficient spectrum bandwidth on AoI.

\section{Conclusion} \label{section:conclusion}
In this paper, we have delved into age-based scheduling for real-time applications in MEC systems. 
Initially, we have refined the concept of AoI to accommodate the event-driven sampling policy and additional processing time. 
Subsequently, we have formulated the problem of AoI minimization and transformed it into an equivalent CMDP. 
We have demonstrated that CMDP can be effectively tackled using RL methods and introduced the novel concept of PDS to accelerate RL's learning process. 
In order to address the limitations inherent in tabular RL, we have merged our algorithm with DDPG and proposed a deep PDS learning approach. 
Numerical results underscore the remarkable efficiency of our algorithm, showcasing its superior performance against benchmark methods in diverse scenarios. 
In our future work, we aspire to (i) extend the applicability of the deep PDS learning algorithm to a wider array of scenarios and 
(ii) enhance its performance by refining neural network architecture and training strategies.


\bibliographystyle{IEEEtran}
\bibliography{ref}

\end{document}